%% file: DDFHP.tex
\documentclass[11pt]{article}
\usepackage{amsmath, amsthm}
\usepackage{amsfonts}
\usepackage{amssymb}
\usepackage{graphicx}
\usepackage{epsf}
\usepackage{epsfig}
\usepackage[normalem]{ulem}

\usepackage[T1]{fontenc}
\usepackage[usenames]{xcolor}

\newcommand{\stkout}[1]{\ifmmode\text{\sout{\ensuremath{#1}}}\else\sout{#1}\fi}
\definecolor{ID2}{rgb}{1.0, 0.2, 0.0}
\definecolor{ID}{rgb}{0.0, 0.0, 0.0}
\definecolor{BH}{rgb}{0.0, 0.5, 0.3}

\usepackage{fullpage}

\input{NN-macros}

\begin{document}

\centerline{\bf {\Large Nonlinear Approximation and (Deep) $\Relu$ Networks}}
\vspace{0.5cm}
\centerline{\bf I. Daubechies, R. DeVore, S. Foucart, B. Hanin, and  G. Petrova \footnote{%
   This research was supported by the NSF grants   DMS 15-21067 (RD-GP),   DMS  18-17603 (RD-GP), 
DMS 16-22134 (SF), DMS 16-64803 (SF),
ONR grants
  N00014-17-1-2908 (RD), N00014-16-1-2706 (RD), and the Simons Foundation Math + X
Investigator Award 400837 (ID).}}
\vspace{1cm}
\begin{abstract}
\noindent
This article is concerned with the approximation  and expressive powers  of deep neural networks.  This is   
an active research area currently producing many interesting papers.  
The results most commonly found in the literature prove that neural networks approximate functions with classical smoothness to the same accuracy as classical linear methods
of approximation, e.g.  
approximation by polynomials or by piecewise polynomials on prescribed partitions.  
However, approximation by neural networks depending on $n$ parameters is a form of nonlinear approximation and as such should be compared with other nonlinear methods such as variable knot splines or $n$-term approximation from dictionaries.

The performance of neural networks in targeted applications such as machine learning indicate that they actually possess even greater approximation power than these  traditional methods of nonlinear approximation. The main results of this article prove  that this is indeed the case.  
This is done by exhibiting large classes of functions which can be efficiently captured  by neural networks where classical nonlinear methods fall short of the task.  

The present article purposefully limits itself to studying the approximation of univariate functions by ReLU networks.  
Many generalizations to functions of several variables and other activation functions can be envisioned.  
However, even in this  simplest of settings considered here, 
a theory that completely quantifies the approximation power of neural networks is still lacking.
 
\noindent
{\bf AMS subject classification:} 41A25, 41A30, 41A46, 68T99, 82C32, 92B20\\
\noindent {\bf Key Words:} neural networks, rectified linear unit (ReLU), expressiveness, approximation power
\end{abstract}

\section{Introduction}
\label{Introduction}

Neural networks produce structured parametric families of functions that have been studied and used for almost 70 years, going back to the work of Hebb in the late 1940's \cite{hebb1949} and of Rosenblatt in the 1950's \cite{rosenblatt1958perceptron}. 
In the last several years, however, their popularity has surged as they have achieved state-of-the-art performance in a striking variety of machine learning problems, from computer vision \cite{krizhevsky2012imagenet} (e.g. self-driving cars) to natural language processing \cite{wu2016google} (e.g. Google Translate) and to reinforcement learning (e.g. superhuman performance at Go \cite{silver2016mastering,silver2017mastering}). 
Despite these empirical successes, even their proponents agree that neural networks are not yet well-understood and that a rigorous theory of how and why they work could lead to significant practical improvements \cite{BB,LBN}. 

An often cited theoretical feature of neural networks is that they produce universal function approximators \cite{cybenko1989approximation, hornik1989multilayer} in the sense that,
given any continuous target function $f$ and a target accuracy $\epsilon>0$, neural networks with enough judiciously chosen parameters give an approximation to $f$ within an error of size $\epsilon$. 
Their universal approximation capacity has been known since the $1980$'s, yet it is not the main reason why neural networks are so effective in practice. 
Indeed, many other families of functions are universal function approximators. 
For example, one can approximate a fixed univariate real-valued continuous target function $f:[0,1]\rightarrow \R$ using Fourier expansions, wavelets, orthogonal polynomials, etc.  \cite{DL}.  
All of these approximation methods are universal. 
Not only that, but in these more traditional settings, through the core results of Approximation Theory \cite{DL,DNL}, 
we have a complete understanding of the properties of the target function $f$ which determine how well it can be approximated given a budget for the number of parameters to be used.  
Such characterizations do not exist for neural network approximation, even in the simplest setting when the target function is univariate and 
the network's activation function is the {\bf Re}ctified {\bf L}inear {\bf U}nit ($\Relu$).

The neural networks used in modern machine learning are distinguished from those popular in the $1980$'s$/90$'s by an emphasis on using \textit{deep} networks (as opposed to shallow networks with one hidden layer). 
If the universal approximation property were key to the impressive recent successes of neural networks, then the depth of the network would not matter since both shallow and deep networks are universal function approximators.

 The present article focuses on the advantages of deep versus shallow  architectures in  neural networks.   Our goal is to put mathematical rigor into  the empirical observation that deep networks can approximate many interesting functions more efficiently, per parameter, than shallow networks (see \cite{HS,telgarsky2015representation,Y1,Y2} for a selection of rigorous results).

In recent years, there has been a number of interesting papers that address the approximation properties of deep neural networks. 
Most of them treat ReLU networks since the rectified linear unit is the activation function of preference in many applications, particularly for problems in computer vision. 
Let us mention, as a short list, some papers most related to our work. 
It is shown in \cite{E} that deep ReLU networks can approximate functions of $d$ variables as well as linear approximation by algebraic polynomials with a comparable number of parameters. 
This is done by using the fact (proved by Yarotsky \cite{Y1}) that power functions $x^\nu$ can be approximated with exponential efficiency by deep ReLU networks.  
Yarotsky also showed that certain classes of classical smoothness (Lipschitz spaces) can be approximated with rates slightly  better than that of classical linear methods {(see \cite{Y2}}).  
The main advantage of deep neural networks is that they can output compositions of functions cheaply. 
This fact has been exploited by many authors (see e.g. \cite{PM}, where this approach is formalized, and \cite{BGKP}  where this property is used to compare deep network approximation with nonlinear shearlet approximation).

In the present paper, we address the approximation power of $\Relu$ networks and, in particular, whether such networks are truly more powerful in approximation efficiency than the classical methods of approximation. 
Although most of our results generalize to the approximation of multivariate functions, we discuss only the univariate setting since this gives us the best chance for definitive results.
Our main focus is on the advantages of depth, i.e., what  advantages are present in deep networks that do not appear in shallow networks. 
We restrict ourselves to ReLU networks since they have the simplest structure and should be easiest to understand.
 
We emphasize that, when discussing approximation efficiency, we assume that 
$f$ is fully accessible and we ask how well $f$ can be approximated by a neural network with $n$ parameters. 
This is in contrast to problems of data fitting 
%using neural networks 
where,
instead of full access to $f$,
we only have some data observations about it.
In the latter case, the approximation can only use the given data and its performance would depend on the amount and form of that data.  
Performance in data fitting is often formulated in a stochastic setting in which it is assumed that the data is randomly generated and both the observations  and the gradient descent parameter updates are noisy. 
The data fitting problem, using a specific form of approximation like neural networks, has two components, commonly referred to as  bias and variance. 
We are concentrating on the bias component. 
It plays a fundamental role not only in data fitting but also in any numerical procedures based on neural network approximation.

Given two integers $W\ge 2$ and $L\ge 1$, we let (precise definitions are given in the next section)
\begin{equation}
\Upsilon^{W,L}:=\{S: \R\gives \R,\,\,S \text{ \ is produced by a }\Relu \text{ network of width }W\text{ and depth }L\},\label{E:Ups-WL-def}
\end{equation}
 and denote by $n(W,L)$ the number of its parameters.
We fix  $W$ and study the approximation families  $\Upsilon^{W,L}$
when the number of layers $L$ is allowed to vary. 
Our interest is in understanding why taking $L$ large, i.e., 
why using  deep networks is beneficial. 
One way to investigate the approximation power of  $\Upsilon^{W,L}$  is to first compare it  to known nonlinear approximation families with essentially the same  number of degrees of freedom. 
Since every element in $\Upsilon^{W,L}$ is a {\bf C}ontinuous {\bf P}iece{\bf w}ise {\bf L}inear (CPwL) function,
the classical approximation family closest to $\Upsilon^{W,L}$ is the nonlinear set
$$
\Sigma_n:=\{ S:\R\gives \R, \,\, S \text{ is a CPwL function with at most }n \text{ 
distinct breakpoints in } (0,1) \}.
$$
The elements of $\Sigma_n$ are also  called free knot linear splines.
We place the restriction that the breakpoints are in $(0,1)$ 
because we are concerned with approximation on the interval $[0,1]$.
 
 When   $n \asymp n(W,L)$, the sets $\Sigma_n$ and $\Upsilon^{W,L}$ have comparable complexity in terms of parameters needed to describe them,
since the elements in $\Sigma_{n}$ are determined by $2n+2$ parameters. 
This comparison also probes the expressive power of depth for $\Relu$ networks because $\Sigma_{W}$ is (essentially) the same as the one-layer $\Relu$ network $\Upsilon^{W,1}$,  see \eqref{comp}.

Several interesting results \cite{daniely2017depth,mehrabi2018bounds,telgarsky2015representation} show that,
 for arbitrarily large $k\geq 1$ and  $n = n(W,L)$ sufficiently large,
\be
\label{biggerset}
\Upsilon^{W,L}\setminus \Sigma_{{n}^k} \neq \emptyset,
\ee
 cf e.g. \cite[Theorem 1.2]{telgarsky2015representation}. 
 This means that sufficiently deep $\Relu$ networks with $n$ parameters can compute certain CPwL functions whose number of breakpoints exceeds any power of $n$ (the increase of the network depth is necessary as $k$ grows). 
 The reason for \eref{biggerset} is that composing two CPwL functions can multiply the number of breakpoints, allowing networks with $L$ layers of width~$W$ to create roughly $W^L$ breakpoints for very special choices of weights and biases. 
 By choosing to use the available $n$ parameters in a deep rather than shallow network, one can thus produce functions with many more breakpoints than parameters, albeit these functions  have a very special structure.

The first natural question to answer in comparing $\Sigma_n$ with $\Upsilon^{W,L}$ is whether,
 for every fixed $W\geq 2$, 
 each function $S\in \Sigma_n$ is in a corresponding set $\Upsilon^{W,L}$ with $n(W,L) \asymp n$, i.e., with a comparable number of parameters.  
 This would guarantee we do not lose anything in terms of expressive power when considering deep networks with fixed width $W$ over shallow networks with fixed depth $L$. 
 One of our results, Theorem \ref{T:main}, gives a resolution to this question and shows that, up to a constant multiplicative factor, fixed-width $\Relu$ networks depending on $n$ parameters are at least as expressive as the free knot linear splines $\Sigma_{n}$.  
 In other words, deep ReLU networks retain all of the approximation power of free knot linear splines but also add something since they can create functions which are far from being in $\Sigma_n$.  
 We want to understand the new functions being created and how they can assist us in approximation and thus in data fitting. In this direction,
 we showcase in \S\ref{sec:selfsimilar} and \S\ref{dictionary} two classes of functions easily produced by $\Relu$ networks,
 one consisting of self-similar functions
 and the other emulating trigonometric functions.
 Appending these classes to $\Sigma_n$ naturally provides a powerful dictionary for nonlinear approximation.

What types of results could effectively explain the increased approximation power of deep networks as compared 
with other forms of approximation?  One possibility is to exhibit classes $K$ of functions on which   the decay rate of approximation error for neural networks is better than for other methods (linear or nonlinear)  while depending on the same number of parameters. 
On this point, let us mention that by now there are several theorems in the literature (see  e.g. \cite{BGKP,CLM,MP,SCC}) which show that neural networks perform as well as certain classical methods such as polynomials, wavelets, shearlets, etc.,  but they do not show that neural networks perform any better than these methods.   

 We seek more convincing results providing compact classes $K$ that are subsets of Banach spaces~$X$ on which neural networks perform significantly better than other methods of approximation. 
 In this direction, we mention at the outset that such sets $K$ cannot be described by classical smoothness (such as Lipschitz, Sobolev, or Besov regularity) because for classical smoothness classes $K$, there are known lower bounds on the performance for any methods of approximation (linear or nonlinear).  These lower bounds are provided by  concepts such as entropy and widths.  However, let us point out that there is an interesting little twist here that allows deep neural networks to give a slight improvement over
 classical approximation methods for certain Lipschitz, Sobolev, and Besov classes  (see  Theorems \ref{Liptheorem} and \ref{Wptheorem}).   
This improvement is possible when the selection of parameters used in the approximation is allowed to be unstable.

Our results on the expressive power of depth describe  certain classes of functions that can be approximated significantly better by $ \Upsilon^{W,L}$ than by $\Sigma_{n}$ when $n(W,L)$ is comparable to $n$, see \S\ref{depth}.   
The construction of these new classes of functions exploits the fact that,
 when $S$ and $T$ are functions in $\Sigma_n$,
   their composition $S\circ T$,
 can be produced by fixed-width $\Relu$ networks depending on a number of parameters comparable to $n$, 
This composition property allows one to construct broad classes of functions, based on self similarity, whose approximation error decays exponentially using deep networks but only polynomially using $\Sigma_n$ (due to the utter failure of this composition property for $\Sigma_n$).

%%%%%%%%%%%%%%%%%%%%%%%%%
%%%%%%%%%%%%%%%%%%%%%%%%%%
%%%%%%%%%%%%%%%%%%%%%%%%%%%%%%%%%

\section{Preliminaries and notation} 
\label{NN-preliminaries-notations}
\begin{figure}[h]
  \centering
\includegraphics[scale=0.83]{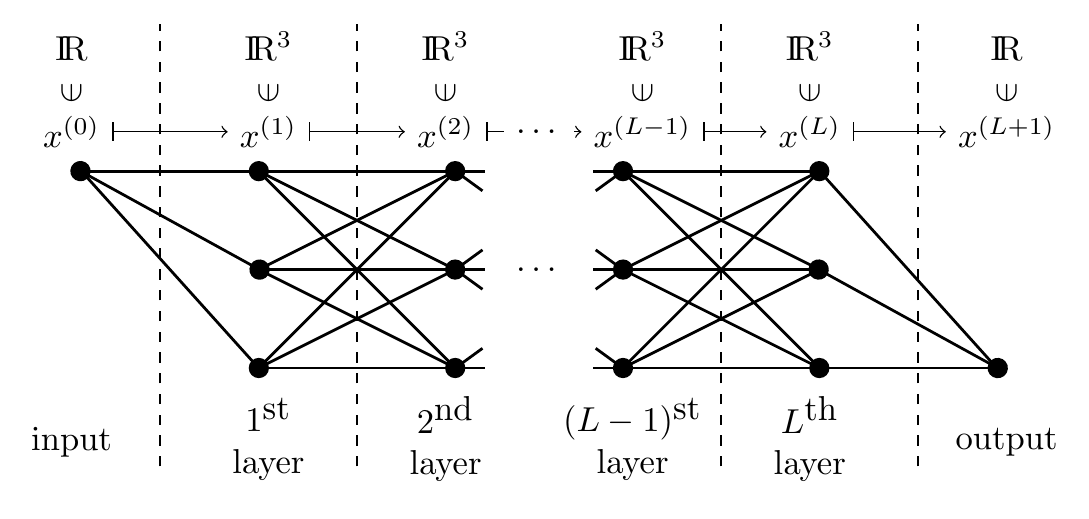}
\caption{The computation graph associated to a neural network with input/output 
dimension $1$,  width $W=3$ and $L$ hidden layers. 
The edges between layers $\ell-1$ and $\ell$ are labeled by the entries of the weight matrix $M^{(\ell-1)}$. 
The $j^{th}$ node (called a neuron) at layer $\ell$ computes the $j^{th}$ component of $x^{(\ell)}$ by taking the dot product of the $j^{th}$ row of $M^{(\ell-1)}$ with the entries of $x^{(\ell-1)}$ and adding it to  the $j^{th}$ entry 
of the vector $b^{(\ell-1)}$ of biases.}
\label{Fig2}
\end{figure}

To set some notation, recall the definition of the ReLU function applied to  $x=(x_1,\dots,x_d)\in\R^d$:
 $$
    \Relu(x_1,\ldots,x_d)=(\Relu(x_1),\ldots,\Relu(x_d))=(\max\set{0,x_1},\ldots,\max\set{0,x_d}).
$$ 
 \begin{definition}
\label{D:net-def}
A fully connected feed-forward $\Relu$ network $\cN$  with  width $W$ and depth $L$  is  a collection of weight matrices $M^{(0)},\dots,$  $M^{(L)}$ and bias vectors $b^{(0)},\dots,  b^{( L)}$.  The matrices $M^{(\ell)}$, $\ell=1,\ldots, L-1$, are of size $W\times W$, whereas  $M^{(0)}$ has size $W\times 1$, and $M^{( L)}$ has size $1\times W$. 
The biases $b^{(\ell)}$ are vectors of size $W$ if $\ell=0,\dots,  L-1$ and a scalar if $\ell=L$. Each such network $\cN$  produces a univariate real-valued function
$$
A^{( L)}\circ \Relu \circ A^{(L-1)}\circ \cdots \circ\Relu \circ A^{(0)}(x),\quad x\in\R,
$$
where 
$$
  A^{(\ell)}(y)=M^{(\ell)}y+b^{(\ell)}, \quad \ell=0,\dots, L.
$$
We define 
$ 
\Upsilon^{W,L}
$ 
as the set of such functions resulting from all possible choices of weights and biases.
  \end{definition}
Every $S \in \Upsilon^{W,L}$ is a CPwL function on the whole real line. 
For each input $x:=x^{(0)}\in \R$, the value $S(x^{(0)})$ of any $S\in \Upsilon^{W,L}$ is computed after the calculation of a series of intermediate vectors $x^{(\ell)}\in \R^W$,   
called vectors of activation at layer $\ell$, 
$\ell=1, \ldots,L$,
before finally producing the output $x^{(L+1)}= M^{(L)}x^{(L)}+b^{(L)}$.  
The computations performed by such a network to produce an $S \in \Upsilon^{W,L}$  are shown schematically in Figure \ref{Fig2}. 
%%%%%%%%%%%%%%%%%%%%%%%%%%%%%%%%%%%%%%%%%%%%%%%%%%%%%%%
%%%%%%%%%%%%%%%%%%%%%%%%%%%%%%%%%%%%%%%%%%%%%%%%%%%%%%%%%

\begin{figure}[h]
  \centering
\includegraphics[scale=.7]{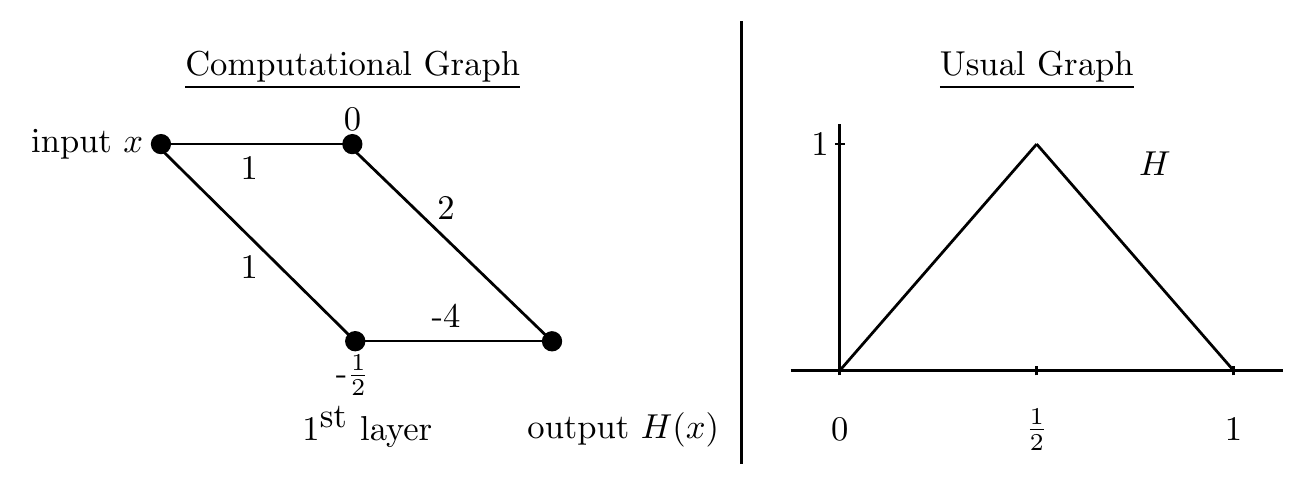}
\caption{The computation graph and usual graph associated to $H$.}
\label{F:hat-function}
\end{figure}

For example, the {\em hat function} (also called {\em triangle function}) $H:[0,1]\rightarrow \R$, defined as
\begin{equation}
\label{hat}
H(x)=2(x-0)_+-4\Big(x-\frac{1}{2}\Big)_+=\begin{bmatrix}     2&-4\end{bmatrix}\Relu\left\{\begin{bmatrix}1 
\\1\end{bmatrix}x+\begin{bmatrix}0 \\-\frac{1}{2}\end{bmatrix}\right\}=
\begin{cases}
2x, \quad 0\leq x\leq \frac{1}{2},\\
2(1-x), \,\,\frac{1}{2}< x\leq 1, 
\end{cases}
\end{equation}
belongs to $\Upsilon^{2,1}$, see Figure \ref{F:hat-function}.
%%%%%%%%%%%%%%%%%%%%%%%%%%
%%%%%%%%%%%%%%%%%%%%%%

 For $L=1$, each function in $\Upsilon^{W,1}$  is a CPwL function with at most $W$ breakpoints determined by the nodes in the first layer.
Conversely, any CPwL function with $(W-1)$ breakpoints interior to  $[0,1]$,  when considered on the interval $[0,1]$, is the restriction of a function from  $\Upsilon^{W,1}$ to that interval.  Indeed, the elements ${\mathcal S}\in \Sigma_{W-1}$ on $[0,1]$ can be represented as 
\begin{eqnarray}
%\label{E:ReLU-spline}
\nonumber
ax+b+\sum_{j=1}^{W-1}m_j(x-\xi_j)_+=
 \begin{bmatrix}     a&m_1&\ldots&m_{W-1}\end{bmatrix}\Relu
\left\{\begin{bmatrix}1\\1\\ \ldots\\1\end{bmatrix}x + \begin{bmatrix}0\\
-\xi_1\\\ldots \\ -\xi_{W-1}\end{bmatrix}\right\}+b,
\end{eqnarray}
where $\xi_1,\ldots,\xi_{W-1}$ are the  interior breakpoints.
In other words,  as functions on $[0,1]$, we have
\begin{equation}
\label{comp}
\Sigma_{W-1} \subset \Upsilon^{W,1} \subset \Sigma_{W},
\end{equation}
which means that,  for large $W$, 
the sets $\Upsilon^{W,1}$ and $\Sigma_W$ are essentially the same. 
Therefore, neural networks with one hidden layer have the same approximation power as CPwL functions with the same number of parameters.

The number of parameters used to generate  functions  in $\Upsilon^{W,L}$ is
\begin{equation}
\label{numberparameters}
n(W,L)~=~W(W+1)L-(W-1)^2+2.
\end{equation}
Not all counted parameters 
(the weights, i.e., entries of $M^{(\ell)}$, and biases, i.e., entries of $b^{(\ell)}$)
are independent, since for instance some of the multipliers used in the transition $x^{(L)}\gives x^{(L+1)}$ could have been absorbed in the preceding layer.
We write
$$
n(W,L) \asymp W^2 L
$$
to indicate that $n(W,L)$ is comparable to $W^2L$,
in the sense that there are constants $c,C>0$ such that
$c \ W^2L \le n(W,L) \le C \ W^2 L$
--- one could take $c=1/2$ and $C=2$ when $W\ge 2$ and $L \ge 2$.

%%%%%%%%%%%%%%%%%%%%%%%%%%%%
%%%%%%%%%%%%%%%%%%%%%%%%%%%%
%%%%%%%%%%%%%%%%%%%%%%%%%%%%
\section{ReLU networks are at least as  expressive as  free knot linear splines} 
\label{sec:Up}

 In this section, we fix $W\geq 4, L\geq 2$, and consider the set $\Upsilon^{W,L}$ defined in \eqref{E:Ups-WL-def}. Our goal is to prove that $\Sigma_n\subset\Upsilon^{W,L}$, where the number of its parameters  $n(W,L)\leq Cn$ for a certain fixed constant $C$.  In order to formulate   our exact result we define  $q:=\lfloor \frac{W-2}{6}\rfloor$ when $W\geq 8$ and $q:=2$ for $4\leq W<7$.

\begin{theorem}
\label{T:main}
Fix a width  $W\ge 4$.    For every $n\ge 1$, the set  $\Sigma_n$ of free knot linear splines with $n$ breakpoints is contained in the set $\Upsilon^{W,L}$ of functions produced by width-$W$ and depth-$L$ ReLU networks,  where
\be
\nonumber 
 L=  
  \begin{cases}
   2\left \lceil \frac{n}{q(W-2)}\right\rceil, \quad 
 & n\geq q(W-2), \\
    2, & n< q(W-2),
  \end{cases}
\ee
\be
\nonumber 
 n(W,L)\leq
  \begin{cases}
   Cn, \quad 
 & n\geq q(W-2), \\
    W^2+4W+1, & n< q(W-2),
  \end{cases}
\ee
with $C$ an absolute constant.
\end{theorem}

Before giving the proof of Theorem \ref{T:main} in \S \ref{NN-sketch-proof} below we first introduce in \S \ref{standard} some notation. 

%%%%%%%%%%%%%%%%%%
%%%%%%%%%%%%%%%%%%%
%%%%%%%%%%%%%%%%%%
\subsection{Special $\Relu$ neural networks}
\label{standard}
\begin{figure}[ht]
  \centering
\includegraphics[width=.8\textwidth]{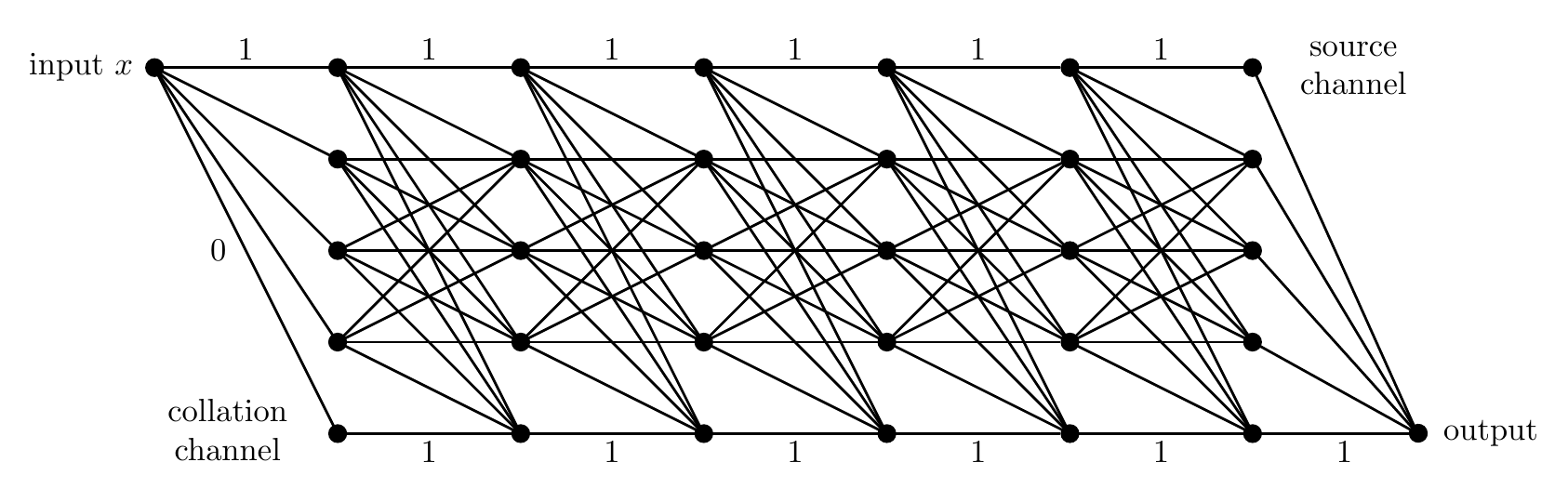}
%\vspace*{0.5 cm}
\caption{The computation graph associated to  $\Up^{5,6}$.}
\label{F:skip_collation_net}
\end{figure}

 Our main vehicle for proving Theorem \ref{T:main}  is a special subset $\Up^{W,L}\subset \Upsilon^{W,L}$, which  we now describe. Given a width  $W\ge 4$ and a depth  $L\ge 2$,  we focus on  networks where a special role is reserved for two nodes in each hidden layer, see Figure \ref{F:skip_collation_net},  which depicts these nodes  as the first (``top'')  and at the last (``bottom'') node of each hidden layer, respectively. 
The top neuron (first node), which is $\Relu$ free, is used to simply copy the input $x$. The concatenation of all these top nodes can be viewed as a special ``channel'' (a term borrowed from the electrical engineering filter-bank literature) that skips computation altogether and just carries $x$ forward. 
We call this the {\em source channel}   (SC). 
The bottom neuron (last node) in each layer, which is also $\Relu$ free, is used to collect intermediate results. We call the concatenation of all these bottom nodes the {\em collation channel} (CC). 
This channel never feeds forward into subsequent calculations, it only accepts previous calculations. The rest of the channels are {\it computational channels} (CmC). 
The fact that  a special role is reserved for two channels enforces the natural restriction $W\geq 4$, since we need at least two computational channels. 
We call these networks (with SC and CC) {\it special} neural networks, for which we introduce a special notation, featuring a top and a bottom horizontal line to represent the SC and CC, respectively. 
Namely, we set 
$$
\Up^{W,L}=\{S: [0,1] \gives \R,\,\,S \text{ \ is produced  by a special
network of width }W\text{ and depth }L\}.
$$
We feel that these more structured networks are not only useful in proving results on approximation but may be useful in applications such as data fitting.
In practice, the designation of the first row as a SC and the last row as a CC amounts to having matrices $M^{(\ell)}$ and vectors $b^{(\ell)}$ of the form
$$
M^{(0)}=\begin{bmatrix}1&m_2^{(0)}&\ldots&m_{W-1}^{(0)}&0\end{bmatrix}^{\top}, \quad
b^{(0)}=\begin{bmatrix}0&b_2^{(0)}&\ldots&b_{W-1}^{(0)}&0\end{bmatrix}^{\top},
$$
\begin{equation}
\label{matrixUpsilon}
M^{\ell)}=\begin{bmatrix}1&0&\ldots&0&0 \\
                                    m_{2,1}^{(\ell)}&m_{2,2}^{(\ell)}&\ldots&m_{2,W-1}^{(\ell)}&0\\
m_{3,1}^{(\ell)}&m_{3,2}^{(\ell)}&\ldots&m_{3,W-1}^{(\ell)}&0\\
\ldots&\ldots&\\
m_{W-1,1}^{(\ell)}&m_{W-1,2}^{(\ell)}&\ldots&m_{W-1,W-1}^{(\ell)}&0 \\
m_{W,1}^{(\ell)}&m_{W,2}^{(\ell)}&\ldots&m_{W,W-1}^{(\ell)}&1
\end{bmatrix}, \quad 
b^{(\ell)}=\begin{bmatrix}0\\b_{2}^{(\ell)}\\b_{3}^{(\ell)}\\\ldots\\b_{W-1}^{(\ell)}\\b_{W}^{(\ell)}\end{bmatrix},\quad 
 \ell=1,\ldots,L-1,
\end{equation}
and 
$$
M^{(L)}=\begin{bmatrix}m_1^{(L)}&\ldots&m_{W-1}^{(L)}&1\end{bmatrix}, \quad b^{(L)} \in \R.
$$

\begin{remark}
\label{relufree}
Note that since the SC and CC are ReLU-free, the width-$W$ depth-$L$ special networks do not form a subset of the set of  width-$W$ depth-$L$ ReLU networks. However, in terms of sets of functions produced by these networks, the inclusion
\begin{equation}
\label{incl}
\Up^{W,L}\subset \Upsilon^{W,L}
\end{equation}
is valid. 
Indeed, given $ \bar S \in \Up^{W,L}$, determined by the set of matrices and vectors $\{\bar M^{(\ell)},\bar b^{(\ell)}\}$, $\ell=0,\ldots,L$, we will construct  $\{ M^{(\ell)},b^{(\ell)}\}$, $\ell=0,\ldots,L$, such that $\bar S$ is also the output of a $\Relu$ network  with the latter matrices and vectors. 
First, notice that the input $x\in [0,1]$, and therefore we have $x= \Relu(x)$. 
Next, since the bottom neuron in the $\ell$-th layer, $\ell=1,\ldots,L$,  collects  a function $ \bar S^{(\ell)}(x)$ depending continuously on $x \in [0,1]$, there is a constant $C_\ell$ such that $ \bar S^{(\ell)}(x) + C_\ell \ge 0$ for all $x \in [0,1]$. Hence $\bar S^{(\ell)}(x) = \Relu( \bar S^{(\ell)}(x) + C_\ell) - C_\ell$. Therefore, the $\Relu$  network that produces $\bar S$ has the same matrices $M^{(\ell)}=\bar M^{(\ell)}$ and vectors $b^{(\ell)}$, $\ell=1,\ldots,L-1$, where
$$
b^{(\ell)}_j=\bar b^{(\ell)}_j, \quad j=1, \ldots,W-1,\quad b^{(\ell)}_W=\bar b^{(\ell)}_W+C_\ell,
$$  
and $b^{(L)}=\bar b^{(L)}-\sum_{\ell=1}^{L-1}C_\ell$.
\end{remark}

\begin{figure}[ht]
  \centering
\includegraphics[width=0.8\textwidth]{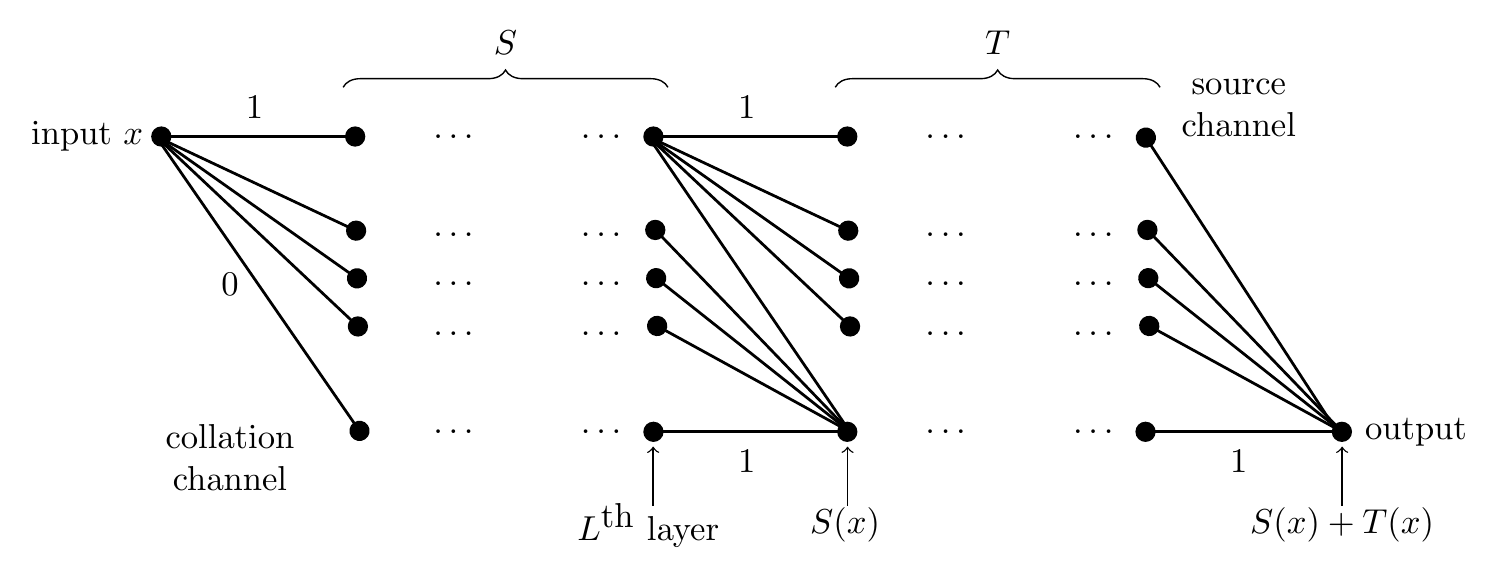}
%\vspace*{0.5 cm}
\caption{The computational graph for summation.}
\label{sum}
\end{figure}

\begin{proposition}
\label{prnew}
Special $\Relu$ neural networks produce sets of CPwL functions  that satisfy the following properties:\\
 {\rm \bf (i)}  For all $W,L,Q$,
\be
\label{NNproperties}
\Up^{W,L}+\Up^{W,Q}\subset\Up^{W,L+Q}.
\ee

{\rm \bf (ii)} For $L<P$, 
$$\Up^{W,L}\subset\Up^{W,P}.$$

\end{proposition}
{\bf Proof:} To show {\bf (i)},
we first fix $S\in \Up^{W,L}$ and $T\in\Up^{W,Q}$ and use the following \lq concatenation\rq \   of the special networks for $S$ and $T$. The concatenated network has the same input and first $L$ hidden layers as the network that produced $S$. Its  $(L+1)$-st layer is the  same as  the first hidden layer of the network that produced $T$ except that in the collation channel it places $S$ rather than $0$. The remainder of the concatenated network is the same as the remaining layers  of the network producing 
$T$ except that the collation channel is updated, see Figure \ref{sum}. The proof of  {\bf (ii)}  follows the proof of   (i) with $Q=P-L$ and $T\equiv 0$. 
\hfill $\Box$
%%%%%%%%%%%%%%%%%%
%%%%%%%%%%%%%%%%%%%
%%%%%%%%%%%%%%
%%%%%%%%%%%%%%%

%%%%%%%%%%%%%%%%%%%%%%%%%%
%%%%%%%%%%%%%%%%%%%%%%%%%%%
%%%%%%%%%%%%%%%%%%%%

\subsection{Proof of Theorem \ref{T:main}}
\label{NN-sketch-proof}

In this section, we prove Theorem \ref{T:main}. Namely, we show that for any fixed width $W\ge 4$, any element  $T$ in $\Sigma_n$ is the output of a special network with a number of parameters comparable to~$n$.
 
Our constructive proof  begins with Lemma \ref{core}, in which we create a special $\Relu$ network with only $2$ layers that generates a particular collection of CPwL functions, see \eref{Sdef}. To describe this collection, we consider any positive integer $N$ of the form $N:={q(W-2)}$, where  $q:=\lfloor (W-2)/6\rfloor$. Since it is meaningful to have only cases when $q\geq 1$, we impose the restriction $W\geq 8$. In the Appendix, we treat the remaining cases when $4\le W<8$. Notice that $N$ is small and so at this stage we are only showing how to construct CPwL functions with a few breakpoints.

\begin{figure}[ht]
  \centering
\includegraphics[width=.8\textwidth]{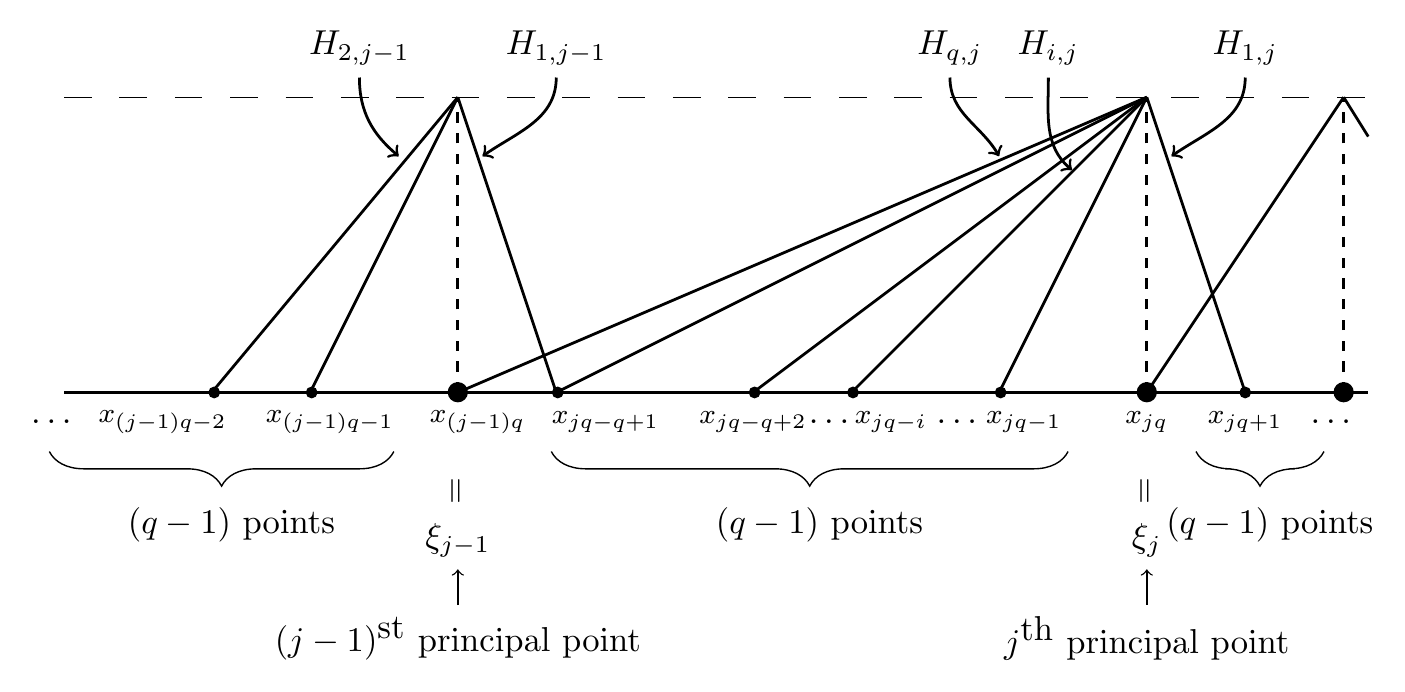}
%\vspace*{0.8 cm}
\caption{The graphs of $H_{i,j}$.}
\label{H}
\end{figure}

Let $ x_1<\cdots<x_{N}\in (0,1)$ be any $N$ given  breakpoints in $(0,1)$ 
and choose  $x_0$ and  $x_{N+1}$ to be any two additional points such that $0\leq x_0<x_1$ and $1\geq x_{N+1}>x_N$. The set of all  CPwL functions which vanish outside of $[x_0,x_{N+1}]$ and have breakpoints only at the  $x_0, x _1,\dots, ,x_N ,x_{N+1}$ is denoted by
\begin{equation}
\label{Sdef}
\cS:=\cS(x_0,\dots,x_{N+1})
\end{equation}
and  is a linear space of dimension $N$. We create a basis for $\cS$ the following way.
We denote by $\xi_j$, $j=1,\dots,(W-2)$, the  points $\xi_j:= x_{jq}$, which we call  principal breakpoints and to each principal breakpoint $\xi_j$, we associate $q$ basis functions  $H_{i,j}$, $i=1,\dots,q$. Here $H_{i,j}$, see Figure \ref{H}, 
%%%%%%%%%%%%%%%%%%
%%%%%%%%%%%%%%%%%%%
%%%%%%%%%%%%%%
%%%%%%%%%%%%%%%
is a hat function supported on $I_{i,j}:= [x_{jq-i,}, x_{jq+1}]$ which takes the value 
$0$ at the endpoints of this interval, the value one at $\xi_j$ and is linear
  on each of the two intervals $[x_{jq-i,}, x_{jq}]$ and $[x_{jq}, x_{jq+1}]$, that is
\begin{eqnarray*}
H_{i,j}(x)&=& \begin{cases}\frac{x-x_{jq-i}}{x_{jq}-x_{jq-i}}, &\mbox{ if }x \in (x_{jq-i}, x_{jq}),\\ 
0, &\mbox{ if}  \ x\notin I_{i,j},\\
\frac{x-x_{jq+1}}{x_{jq}-x_{jq+1}}, &\mbox{ if }x \in (x_{jq}, x_{jq+1}).
\end{cases}\\
\end{eqnarray*}
We rename  these hat functions as $\phi_k$, $k=1,\dots,N$, and order them in such a way that  $\phi_k$ has leftmost breakpoint $x_{k-1}$. We say $\phi_k$ is associated with $\xi_j$ if $\xi_j$ is the principal  breakpoint where it is nonzero. We claim that  these $\phi_k$'s are a basis for $\cS$.  Indeed, since there are $N$ of them, we need only check that they are linearly independent.  If $\sum_{k=1}^Nc_k\phi_k=0$, then $c_1=0$ because $\phi_1$ is the only one of these functions which is nonzero on $[x_0,x_1]$.   We then move from left to right getting that each coefficient $c_k$ is zero.
 \begin{lemma}
 \label{core}
 For any $N$ breakpoints $x_1<\cdots<x_N\in (0,1)$, $N:={q(W-2)}$, $q:=\lfloor (W-2)/6\rfloor$,  $W\geq 8$,  
$\cS(x_0,\dots,x_{N+1})\subset \Up^{W,2}$.
 \end{lemma}
 
 \noindent
 {\bf Proof:}  Consider $T\in\cS(x_0,\dots,x_{N+1})$, $T=\sum_{k=1}^N c_k\phi_k$,
and determine its principal breakpoints $\xi_1,\ldots,\xi_{W-2}$ (every $q$-th point from the sequence 
$(x_1,x_2,\ldots,x_N)$ is a principal breakpoint).
We next represent the set of indices $\Lambda=\{1,\dots,N\}$ as a disjoint union of $K\le 6q\leq W-2$ sets $\Lambda_i$,
$$
\Lambda=\cup_{i=1}^K\Lambda_i,
$$
where the $\Lambda_i$'s  have the following two properties:
\begin{itemize}
\item  for any   $\Lambda'\in\{\Lambda_1,\dots,\Lambda_{K}\}$, all of the coefficients $c_k$ with $k\in\Lambda'$ of $T$ have the same sign. 
\item  if 
 $k,k'\in\Lambda'$, then the principal  breakpoints $\xi_j$ and $\xi_{j'}$ associated to $\phi_k,\phi_{k'}$ respectively, satisfy the separation property
 $ |j-j'|\ge 3$.
\end{itemize}
We can find
 such a partition as follows.   First, we divide $\Lambda=\Lambda_+\cup\Lambda_-$ where for each $i\in\Lambda_+$, we have $c_i\ge 0$ and for each $i\in \Lambda_-$, we have
 $c_i<0$.  We then divide each of  $\Lambda_+$ and $\Lambda_-$  into  at most  $3q$ sets  having the desired separation property. If  $K<W-2$, we set $\Lambda_{K+1}=\ldots=\Lambda_{W-2}=\emptyset$. 
It  may also  happen that some of the $\Lambda_k$'s, $k\leq K$, are empty. In all cases for which 
$\Lambda_k=\emptyset$,  we set  $T_k=0$, and write
\be
 \label{decomp}
 T=\sum_{k=1}^{W-2} T_k,\quad T_k:=\sum_{i\in \Lambda_k}c_i\phi_i,\quad k=1,\dots, W-2.
 \ee
 Notice that the $\phi_i$, $i\in \Lambda_k\neq \emptyset$, have disjoint supports and so $c_i=T_k(\xi_j)$ where $\xi_j$ is the principal breakpoint associated to $\phi_i$. 

\begin{figure}[ht]
  \centering
\includegraphics[width=.6\textwidth]{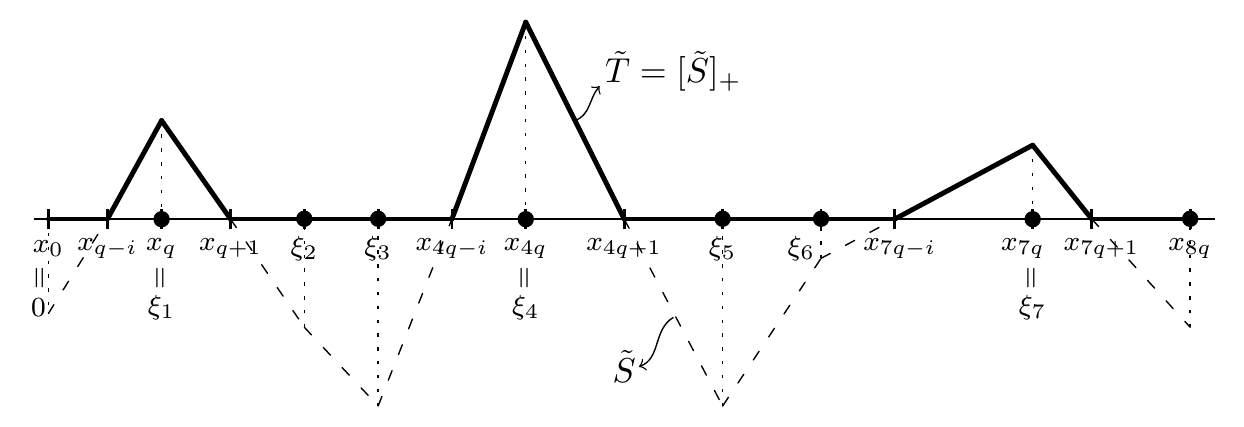}
%\vspace*{0.8 cm}
\caption{A typical $\tilde T$ computed by a node in the second layer of $\Up^{W,2}$.}
\label{Sgraph}
\end{figure}

 We next show that each of the $T_k$ corresponding to a nonempty $\Lambda_k$ is of the form 
$\pm [S_k(x)]_+$ for some linear combination $S_k$  of the $(x-\xi_j)_+$.  
 Fix $k$ and first consider the case  where all of the $c_i$ in $\Lambda_k$ are nonnegative.  We consider the CPwL function   $S_k$ which takes the value $c_i$ at each principal
 breakpoint  $\xi_j$  associated to an $i\in \Lambda_k$.   At the remaining principal breakpoints, we assign negative values to the $S_k(\xi_j)$'s.  We choose these negative values so
 that for any $i \in \Lambda_k$, $S$ vanishes at the leftmost and rightmost breakpoints of all $\phi_i$
with $i\in \Lambda_k$.  
This is possible because of the separation property (see the appendix for a particular strategy for defining the $\Lambda_k$).   It follows that
 $[S_k(x)]_+=T_k(x)$.  A similar construction applies when all the coefficients in $\Lambda_k$ are negative.   In this case, $T_k=-[S_k]_+$ for the constructed $S_k$.
 A typical $T_k$, which for the sake of simplicity we call $\Tilde  T$,  and its decomposition is pictured in Figure \ref{Sgraph}, see \S \ref{ss1}.
  
%%%%%%%%%%%%%%%%%%%%
%%%%%%%%%%%%%%

We can now describe  the  $\Relu$ network that generates $T$.  Since it is special, we focus  only the computational channels. The computational nodes 
in the first layer are $(x-\xi_j)_+$, $j=1,\dots,W-2$, where the $\xi_j$'s are the principal breakpoints.
 The computational nodes in the second layer  are equal to the 
$[S_k]_+$ or $0$.  Because of \iref{decomp}, the target $T$ is the output of this network with output 
layer weights $\pm 1$ or $0$.   
\hfill $\Box$

\vskip .2in
 
\begin{remark}
\label{notenough}
If we want to generate with the same special $\Relu$ network all spaces
$\cS(x_0,\ldots,x_{N_0+1})$ with  $N_0<N$, we can artificially add $(N-N_0)$ distinct points  in the interval
$(x_{N_0},x_{N_0+1})$ and view the elements in $\cS(x_0,\ldots,x_{N_0+1})$ as CPwL with $N$ breakpoints
vanishing outside $[x_0,x_{N_0+1}]$, even though the last $N-N_0+1$ points are not really a breakpoints, except 
possibly $x_{N_0+1}$. 
\end{remark}

%%%%%%%%%%%%%%%%%%%
\begin{figure}[ht]
  \centering
\includegraphics[width=.8\textwidth]{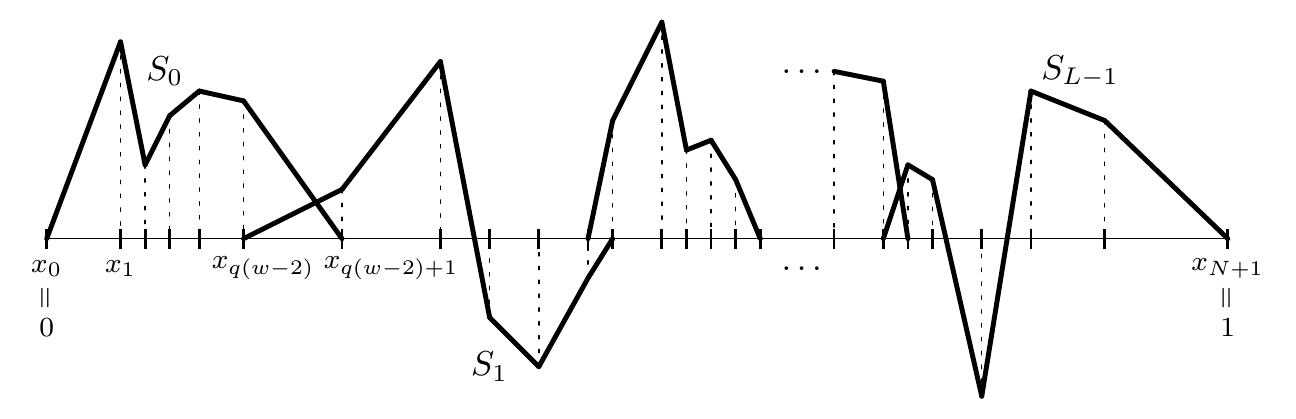}
%\vspace*{0.5 cm}
\caption{The graphs of $S_j$, $j=0,\ldots,L-1$.}
\label{bits}
\end{figure}
%%%%%%%%%%%%%%
%%%%%%%%%%%%%%%
Our next lemma shows how to  carve up the target function $T\in \Sigma_{n}$ with a 
(possibly) large number of breakpoints into ``bitesize'' pieces that are handled by Lemma \ref{core}.
  \begin{lemma}
 \label{lemma2}
 If $T\in \Sigma_N$ is any CPwL function on $[0,1]$ with $N=q(W-2)L$, $q:=\lfloor \frac{W-2}{6}\rfloor$,
$W\geq 8$, 
then $T$ is the output of a special $\Relu$ network  $\Up^{W,2L}$ with  at most $2L$ layers.
  \end{lemma}
  
  \noindent
  {\bf Proof:}  Let $x_1<\cdots<x_N$ be the breakpoints of $T$ in $(0,1)$ and  set  $x_0:=0, x_{N+1}:=1$.  
We define  $\ell(x):=ax+b$ to be the linear function which interpolates $T$ at the endpoints $0,1$ and set  $S:=T-\ell$.  
We can write  $S=S_0+\dots +S_{L-1}$, 
where $S_j\in \Sigma_N$ is the CPwL function which agrees with $S$ at the points $x_i$,   for all indices
$i\in \{ jq(W-2)+1,\dots, (j+1)q(W-2)\}$ and is zero at all other breakpoints of $T$, see Figure \ref{bits}.

\begin{figure}[ht]
  \centering
\includegraphics[width=1.0\textwidth]{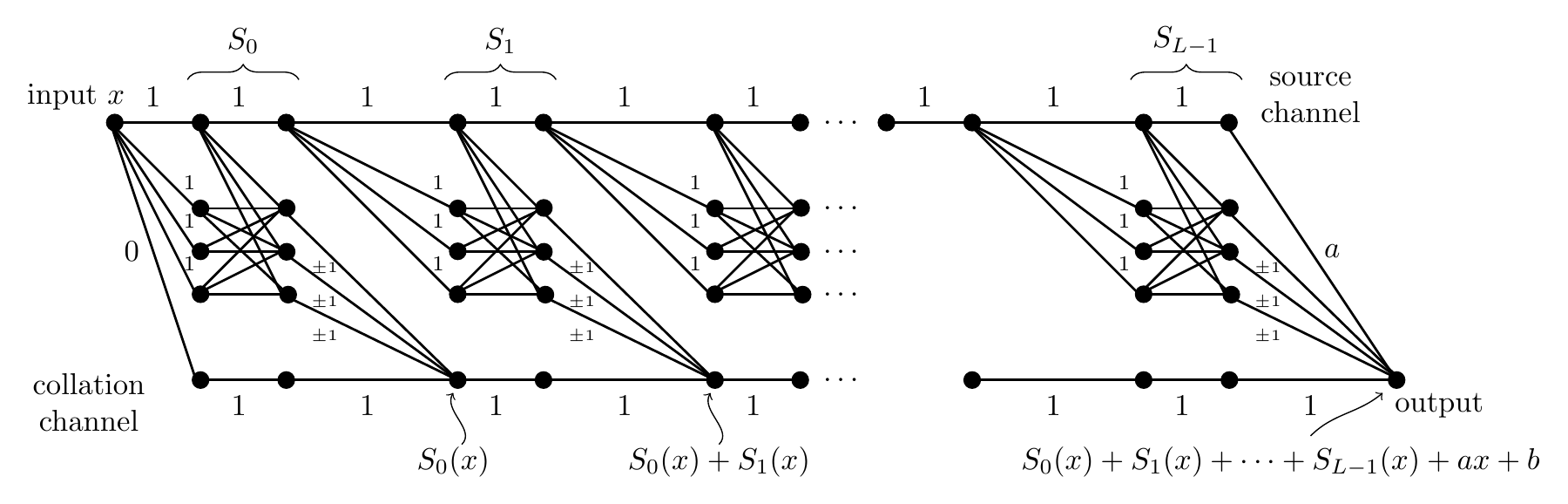}
%\vspace*{0.5 cm}
\caption{The resulting network with $2L$ layers.}
\label{2L}
\end{figure}

Clearly, see \eref{Sdef},
$$
S_j\in \cS(x_{jq(W-2)},\ldots,x_{(j+1)q(W-2)+1}), \quad  j=0,\ldots,L-1,
$$
and therefore, it follows from  Lemma \ref{core} that each $S_j\in\Up_j^{W,2}$. 
We concatenate the $L$ networks that produce $S_j\in\Up_j^{W,2}$, $j=0,\ldots,L-1$, as described in  Proposition \ref{prnew} and thereby produce $S$.   
In order to account for the linear term $\ell(x)$, we assign 
weight $a$  and bias $b$ to the output of the  node of the  skip channel in the last layer of the concatenated network, see Figure \ref{2L}.
%%%%%%%%%%%%%%%%%%
%%%%%%%%%%%%%%%%%%%
%%%%%%%%%%%%%%
%%%%%%%%%%%%%%%
\hfill $\Box$

{\bf Proof of Theorem \ref{T:main}:} Now we are ready to complete the proof of Theorem \ref{T:main}.

\noindent
{\bf Case 1:}  We first consider the case when $W\geq 8$. 
Lemma~\ref{lemma2} and inclusion \eref{incl} show that $\Sigma_N\subset \Upsilon^{W,2L}$ with $N=q(W-2)L$ and $q:=\lfloor \frac{W-2}{6}\rfloor$.
  Given $n$, we choose $N$ as the smallest $N$ of the above form for which $n\le N$,
Let $N_1:= q(W-2)$.   If $n\ge N_1$,
 we choose $L$ as
$$
L~=~L(n,W)~:=~\left\lceil \frac{n}{q(W-2)}\right\rceil,
%\qquad q: = \left\lfloor\frac{W-2}{6}\right\rfloor,
$$
and thus
$
%\frac{n}{q(W-2)}\leq 
L<\frac{n}{q(W-2)}+1.
$
Using \eqref{numberparameters}, we have that the number of parameters in $\Upsilon^{W,2L}$ is
$$
n(W,2L) < 2W(W +1)\left( \frac{n}{q(W-2)}+1\right)-(W-1)^2+2
=\frac{2W(W+1)}{q(W-2)}n+W^2+4W+1.
$$
Optimizing over $W$ show that the maximum of  $\frac{2W(W+1)}{q(W-2)}$ over integers $W\geq 8$ is achieved at $W=13$ and  $q=1$, giving the value $\frac{364}{11}<34$. Hence, 
$$
n(W,2L) <  34 n+W^2+4W+1<34n+ 27q(W-2)\leq  61n,
$$ 
where we used 
that  $W/13\leq q$ and $q(W-2)=N_1\leq n$.

On the other hand if 
 $n< N_1:=q(W-2)$, then
Lemma~\ref{lemma2} and inclusion \eref{incl} show that $\Sigma_n\subset \Sigma_{N_1}\subset \Upsilon^{W,2}$.
Then, we have   
$$
n(W,2)= W^2+4W+1,
$$
as desired.

\noindent
{\bf Case 2:}  The proof of the case $4\leq W<7$ is discussed in the appendix.

%{\bf Case 2:}  The  remaining cases $2\le W-2\leq 5$ which were not covered by our previous analysis are discussed in %the Appendix. {\anew this has to be redone in the Appendix}
\hfill $\Box$

\begin{remark}  We have not tried to optimize constants in the above theorem.  If one counts the actual number of parameters 
used  in $\Upsilon^{W,L}$ (rather than the parameters
available), one obtains   a  much better constant.  We know, in fact, that we can present other  constructions (different than those given here)  which  provide a better constant   in the statement of Theorem {\rm \ref{T:main}}.
\end{remark}

%%%%%%%%%%%%%%%%%%%%
%%%%%%%%%%%%%%%%%%%%%
%%%%%%%%%%%%%%%%%%%%%

\section{More about standard and special networks}
\label{general}
In this section,
we discuss further properties of the sets $\Upsilon^{W,L}$ and $\Up^{W,L}$.
We highlight in particular  Theorem \ref{compos}, which is  a generalization of Theorem \ref{T:main},
and whose proof is deferred to the appendix.
 Note that the conclusion of Theorem \ref{T:main} depends on the ranges of the width $W$ and the parameter  $n$ in $\Sigma_n$.
To avoid excessive notation, we concentrate on only one of these ranges in the theorem below.

\begin{theorem}\label{compos}
The following statement holds for compositions and sums of compositions of 
free knot linear splines:

{\bf (i)} For nonconstant functions $S_1 \in \Sigma_{n_1},\ldots, S_k \in \Sigma_{n_k}$ 
 with $n_i\geq (W-2)\lfloor \frac{W-2}{6}\rfloor$,  and $W\geq 8$,  the composition
\begin{equation}
\label{E:pure-compos}
S_k \circ \cdots \circ S_1 \in \Upsilon^{W,L},
\qquad \quad
L = 2\sum_{j=1}^k\left \lceil \frac{n_j}{\lfloor \frac{W-2}{6}\rfloor(W-2)}\right\rceil,
\end{equation}
where the number of parameters describing $\Upsilon^{W,L}$ satisfies the bound
$$
n(W,L) \le 34\sum_{j=1}^kn_j+2k(W^2+W).
$$

{\bf (ii)} For nonconstant functions $S_{i,j} \in \Sigma_{n_{i,j}}$, $i=1,\ldots, m$, $j=1,\ldots,\ell_i$, 
 with  $n_{i,j}\geq (W-4)\lfloor \frac{W-4}{6}\rfloor$,  and
$W\geq 10$, 
 the sum of compositions satisfies
\begin{equation}
\label{E:full-compos}
\sum_{i=1}^ma_iS_{i,\ell_i} \circ\cdots \circ S_{i,1} 
\in \Up^{W,L}\subset \Upsilon^{W,L},
\end{equation}
where the number of parameters describing  $\Upsilon^{W,L}$ satisfies the unequality
$$
n(W,L) \le 44\sum_{i=1}^m\sum_{j=1}^{\ell_i}n_{i,j}+2W(W+1)\sum_{i=1}^m\ell_i.
$$
\end{theorem}

Theorem \ref{compos} relies on some properties of standard and special networks.
We state and prove below the ones that are explicitly needed in the remainder of paper,
starting with the following results.

\begin{proposition}
\label{composnetsup}
Let $W\geq 2$.
For any ${\cal Y}_1 \in \Upsilon^{W,L_1},\ldots, {\cal Y}_k \in \Upsilon^{W,L_k}$,

{\bf (i)} the composition of the ${\cal Y}_i$ satisfies
\begin{equation}
\label{Up:pure-compos}
{\cal Y}_k \circ \cdots \circ {\cal Y}_1 \in \Upsilon^{W,L},
\qquad
L = L_1 + \cdots + L_k;
\end{equation}

{\bf (ii)} the sum of the ${\cal Y}_i$ satisfies
\begin{equation}
\label{Up:full-compos}
{\cal Y}_1 + \cdots + {\cal Y}_k \in \Up^{W+2,L},
\qquad
L = L_1 + \cdots + L_k;
\end{equation}

{\bf (iii)} the sum of the $({\cal Y}_i)_+ :=  \Relu({\cal Y}_i)$ satisfies
\begin{equation}
\label{Up:full-composrelu}
({\cal Y}_1)_+ + \cdots + ({\cal Y}_k)_+ \in \Up^{W+2,L},
\qquad
L = k+ L_1 + \cdots + L_k.
\end{equation}
\end{proposition}

\begin{figure}[ht]
\label{co}
  \centering
\includegraphics[width=0.7\textwidth]{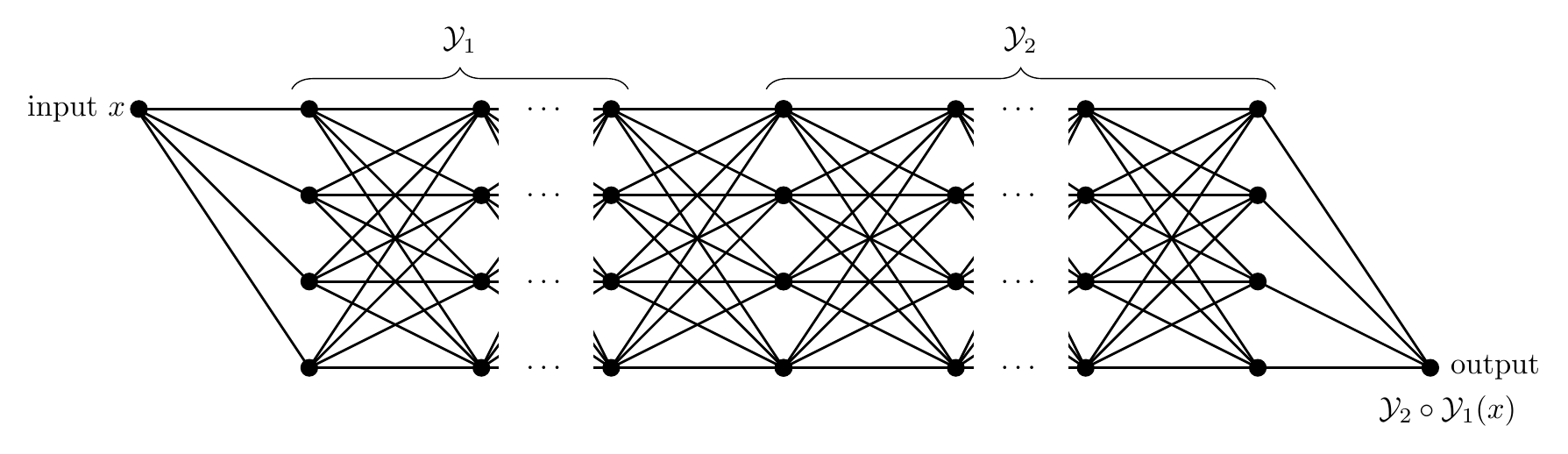}
%\vspace*{0.5 cm}
\caption{The network computing ${\cal Y}_2\circ {\cal Y}_1$.}
\label{co2}
\end{figure}

{\bf Proof:} The argument is constructive. 
First,
to prove \eqref{Up:pure-compos}, 
let ${\cal N}_j$ be the $\Relu$ network with width~$W$ and depth $L_j$ producing ${\cal Y}_j$.
We concatenate the networks ${\cal N}_1,\cdots,{\cal N}_k$ as shown in Figure~\ref{co2} for the case of 
${\cal Y}_2\circ {\cal Y}_1$.
%%%%%%%%%%%%%%%%%
%%%%%%%%%%%%%%%%%%%
The concatenated network has the same input and first $L_1$ hidden layers as the network ${\cal N}_1$. 
Its $(L_1+1)$-st layer is the same as the first hidden layer of the network ${\cal N}_2$. 
The weights between the $L_1$-st  and $(L_1+1)$-st layer are the output weights of ${\cal Y}_1$, 
multiplied by the input weights 
for the first hidden layer of ${\cal Y}_2$. 
%%%%%%%%%%%%%%%%
\begin{figure}[ht]
\label{final}
  \centering
\includegraphics[width=0.7\textwidth]{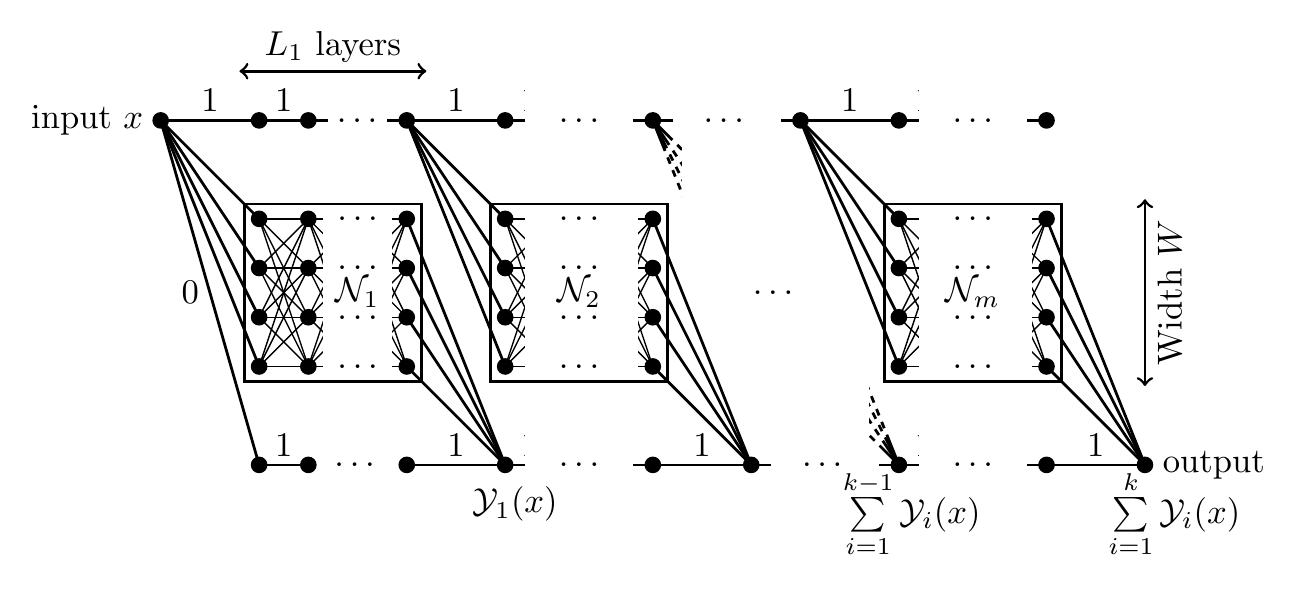}
%\vspace*{0.5 cm}
\caption{The computational graph of the special $\Relu$ network producing  $\sum_{j=1}^k{\cal Y}_j$.}
\label{final}
\end{figure}
%%%%%%%%%%%%%%%%%%%%
The remainder of the concatenated network is the same as the remaining layers of ${\cal N}_2$. 
Clearly, the resulting network will have $n=L_1+\cdots+L_k$ hidden layers.

To show \eqref{Up:full-compos},
we concatenate the networks ${\cal N}_1,\ldots,{\cal N}_k$ as shown in  Figure~\ref{final} by adding a source channel  and a collation channel. The resulting network is a special  network with width $W+2$ and depth 
$L_1+\cdots+L_k$.

Finally, for \eqref{Up:full-composrelu},
we concatenate the networks ${\cal N}_1,\ldots,{\cal N}_k$
by adding  an extra layer after each ${\cal N}_j$ to perform the $\Relu$ operation 
on its output, see Figure~\ref{ccplus}. 
The  rest of the construction is similar to the one for \eqref{Up:full-compos}.
%%%%%%%%%%%%%%%%
\begin{figure}[ht]
\label{ccplus}
  \centering
\includegraphics[width=0.8\textwidth]{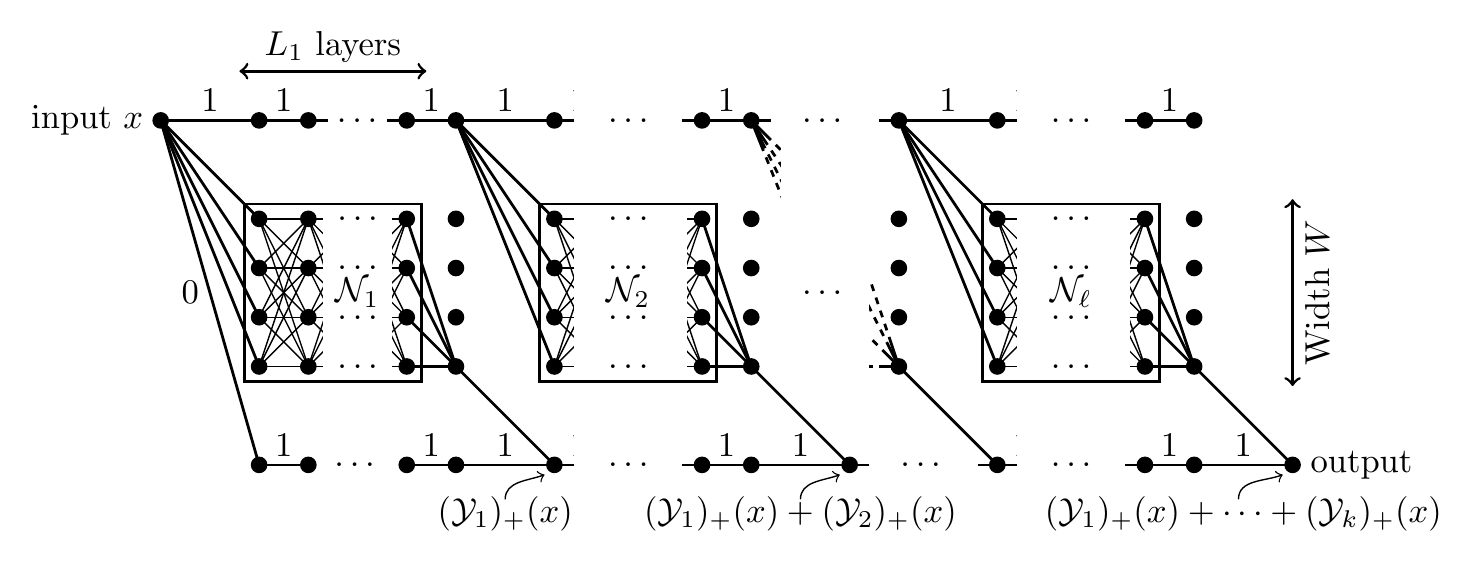}
%\vspace*{0.5 cm}
\caption{The computational graph of the special $\Relu$ network producing $\sum_{j=1}^k({\cal Y}_j)_+$.}
\label{ccplus}
\end{figure}
%%%%%%%%%%%%%%%%%%%%
\hfill $\Box$

%The proof of {\bf (iv)} follows if we concatenate appropriately the networks producing each of 
%$\Upsilon^{W,M_j}$ and $\Upsilon^{W,L_j}$, following the constructions in {\bf (ii)} and {\bf (iii)}.

The following two results will also be needed later.  We use the notation $g^{\circ k}$ to denote the function which results when $g$ is composed with itself $k$ times.

\begin{proposition}
\label{cor1}
 If  $T\in\Upsilon^{{\rm w},L}$, $2\leq {\rm w}\leq W$, then $S= \sum_{i=1}^m a_iT^{\circ i}$ can be produced by a special $\Relu$ network with width $W+2$ and depth $L m$ , that is
 $S \in \Up^{W+2,Lm}$.
\end{proposition}
{\bf Proof:}
First, note that  we have the inclusion
$\Upsilon^{{\rm w},L}\subset \Upsilon^{W,L}$ for every $2\leq {\rm w}\leq W$.
We can always assign  zero weights and biases to any selected   nodes of the network producing 
$\Upsilon^{W,L}$, and therefore we can always assume that $T\in\Upsilon^{W,L}$.
We adjust the network generating $T^{\circ m}$ encountered in the proof of \eqref{Up:pure-compos}.
We augment it to a special network in such a way that,
 after the computation of each of the $T^{\circ i}$, 
 we place $a_iT^{\circ i}(x)$ into the collation channel, see Figure~\ref{co}. 
%%%%%%%%%%%%%%%%
\begin{figure}[ht]
\label{cc}
  \centering
\includegraphics[width=0.8\textwidth]{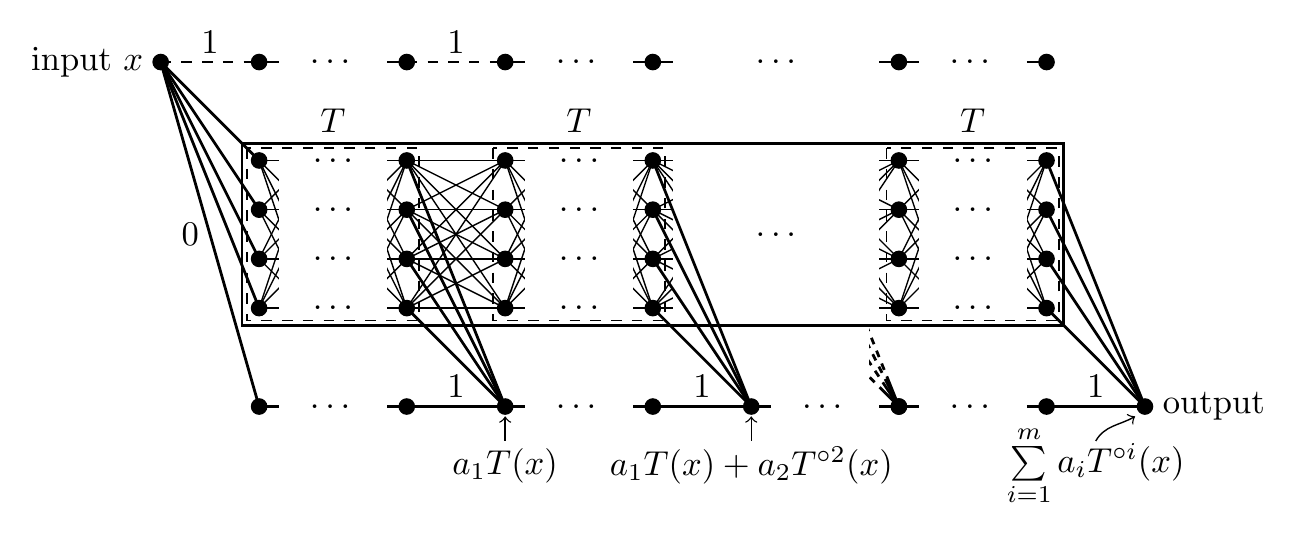}
%\vspace*{0.5 cm}
\caption{The computational graph of the special $\Relu$ network producing  S.}
\label{co}
\end{figure}
%%%%%%%%%%%%%%%%%%%%
The source channel is not needed in this case,
but we include it nonetheless since it will be used when creating the sum of $S$ with another  function.
 \hfill $\Box$
%%%%%%%%%%%%%%%%%%%%%%%%%%%%%%%%%%%%%%%%%

\begin{figure}[ht]
\label{cc}
  \centering
\includegraphics[width=0.8\textwidth]{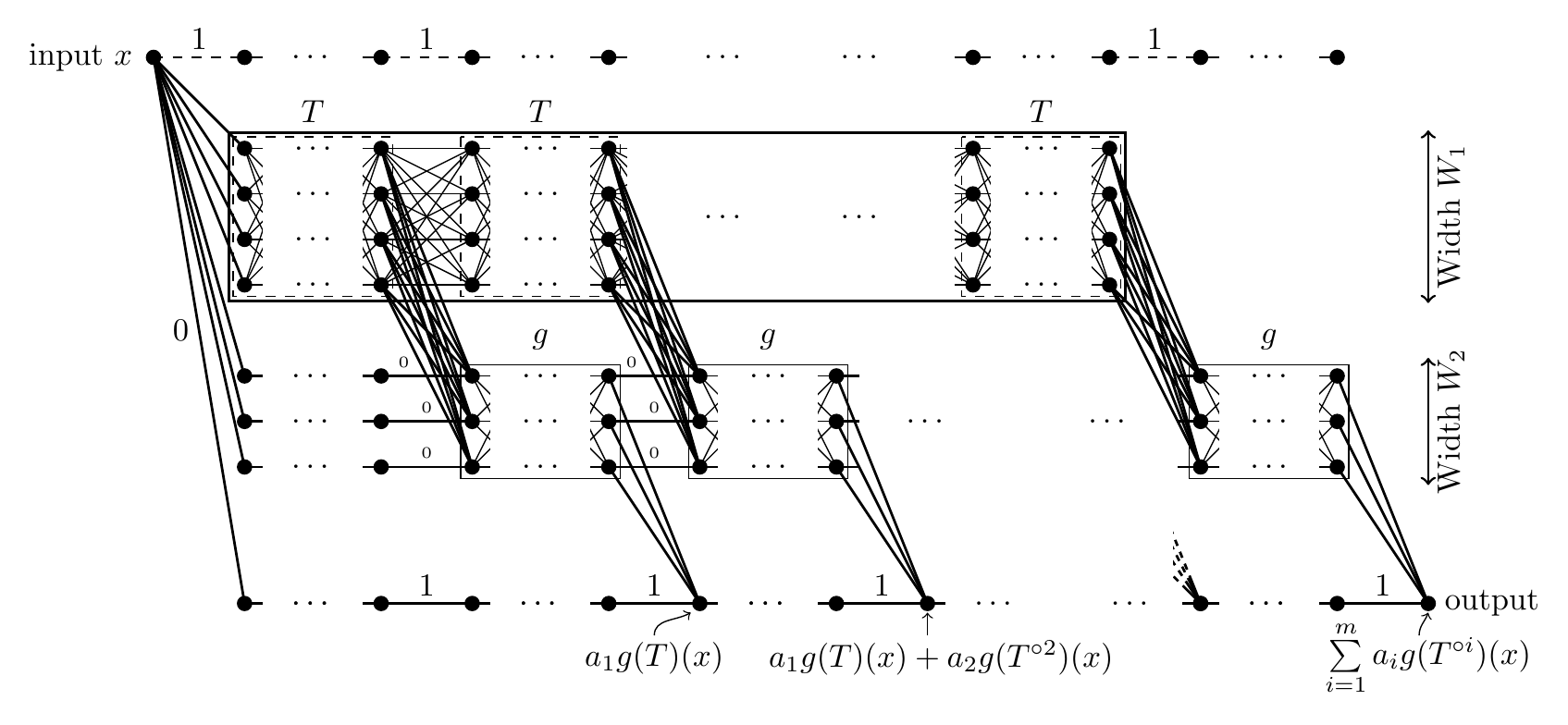}
%\vspace*{0.5 cm}
\caption{The computational graph of the special $\Relu$ network producing  $S_g$.}
\label{2Lg}
\end{figure}

\begin{proposition}
\label{cor2}
 If $T\in\Upsilon^{W_1,\ell}$,  $g\in\Upsilon^{W_2,\ell}$, and $W_1+W_2=W$, then
$S_g=\sum_{i=1}^m a_ig(T^{\circ i})$ can be produced by a special $\Relu$ network with width $W+2$ and depth 
$\ell(m+1)$, i.e., 
$S_g \in  \Up^{W+2,\ell(m+1)}$.
\end{proposition}
{\bf Proof:}
As before,  
we use the network of width $W_1$ generating $T^{\circ m}$.
 For the other $W_2$ channels,
 we use $m$ copies of the network ${\cal G}$ producing $g$
  and combine them as shown in Figure~\ref{2Lg}. 
  After the computation of each of the $T^{\circ i}$, 
  we place $T^{\circ i}(x)$ as an input in the $i$-th copy of  ${\cal G}$ and 
put $a_i$ times its output into the collation channel.
%%%%%%%%%%%%%%%%
%%%%%%%%%%%%%%%%%%%%
Again, the source channel is not needed here but can  be used at a later time.
 \hfill $\Box$
%%%%%%%%%%%%%%%%%%%%%%%%%%%%%%%%%%%%%%%

%%%%%%%%%%%%%%%%%%%
%%%%%%%%%%%%%%%
%%%%%%%%%%%%%%%%%

\section{$\Relu$ networks efficiently  produce functions with self similarity }
\label{sec:selfsimilar} 

Having established that $\Relu$ networks contain sums and compositions of CPwL functions,
we show  that they   also  contain CPwL functions with certain self-similar patterns.  
We formalize this structure below.
  
 Let $0<\xi_1<\xi_2<\cdots<\xi_k<1$ be a fixed set of breakpoints and let 
$S$ be any element of $\cS({\bf \xi}):=\cS(0,\xi_1,\ldots,\xi_k,1)$.  In particular, $S$ vanishes  
outside of $[0,1]$.  We think of $S$ as a {\it pattern}.  It is easy and cheap for $\Relu$ networks
to replicate this pattern on many intervals.   
To describe this,  let $\{J_1,\dots,J_m\}$ denote a collection of $m$ intervals  contained in $[0,1]$ whose interiors are pairwise disjoint.  
 We order these intervals  from left to right.    
 We say that a CPwL function  $F$ is self similar with  pattern
$S \in \cS({\bf \xi})$ if 
\be
\label{fractal}
 F(x)=\sum_{i=1}^m S(h_i(x-a_i)), \quad x\in [0,1],
 \ee
 where $J_i=[a_i,b_i]$ and $h_i=|J_i|^{-1}$,  $i=1,\dots,m$. 
Thus, the function $F$ consists of a dilated version of $S$ on each of the $m$ intervals   $J_i$.
It has roughly $km$ breakpoints but is only described by $2(k+m)$ parameters.
We show below that, 
in order to produce such a function $F$,
 $\Relu$ networks only need a number of parameters of the order $k+m$, 
and not $km$ as  would be naively inferred by regarding $F$ as an element of $\Sigma_{km}$.

\begin{theorem}
\label{fractallemma}
 Let $W \ge 8$.
Any self-similar function $F$ of the form {\rm \eref{fractal}} with 
$S \in \cS({\bf \xi})\subset\Sigma_k$ belongs to $\Up^{W,L}$,  for a suitable value of $L$ that satisfies   $n(W,L) \le C_1(k+m)+C_2W^2$ for some absolute constants
 $C_1,C_2>0$.
\end{theorem}
{\bf Proof:}
 We start with the case when 
$S$ is nonnegative and the intervals $J_i = [a_i,b_i]$ (not just their interiors) are disjoint.
For each $i=1,\ldots, m$,
we introduce a point $c_i$ in the interval $(b_i,a_{i+1})$, where $a_{m+1}:=1$.
We consider the hat function ${\cal H}_i$
which is zero outside $[a_i,c_i]$,
equal to one at $b_i$, and linear on $[a_i,b_i]$ and $[b_i,c_i]$,
as well as the hat function $\hat{\cal H}_i$ which is zero outside $[b_i,a_{i+1}]$,
equal to one at $c_i$, and linear on $[b_i,c_i]$ and $[c_i,a_{i+1}]$.
In the case when $b_m=1$, we cannot construct ${\cal H}_m$ and $\hat{\cal H}_m$ as above, and instead 
set ${\cal H}_m(x)=\frac{1}{1-a_m}(x-a_m)_+$ and $\hat{\cal H}_m(x) = 0$.
With $\hat{S}(x): = S(1-x)$, we claim that 
$$
F = \big( S \circ T - \hat{S} \circ \hat{T} \big)_+,
\qquad \quad \text{where}\quad T := \sum_{i=1}^m {\cal H}_i,
\qquad \hat{T} := \sum_{i=1}^{ m} \hat{\cal H}_i .
$$
This can be easily verified by separating into the three cases $x \in [a_i,b_i]$, $x \in [b_i,c_i]$,
and $x \in [c_i,a_{i+1}]$.
According to Theorem \ref{T:main}, we have $S,\hat{S} \in \Upsilon^{W-4,L'}$
with  either $W^2L'\asymp n(W-4,L')\leq C'k$ or $L'=2$,
and $T,\hat{T} \in \Upsilon^{W-4,L''}$
with  either $W^2L''\asymp n(W-4,L'')\leq C''m$ or $L''=2$.
Then, by Proposition \ref{composnetsup},
we obtain that both 
$S \circ T, \hat{S} \circ \hat{T} \in \Upsilon^{W-4,L'+L''}$,
that their difference $S \circ T - \hat{S} \circ \hat{T} \in \Up^{W-2,2(L'+L'')} \subset \Upsilon^{W-2,2(L'+L'')}$.
At last, the function $F = \big( S \circ T - \hat{S} \circ \hat{T} \big)_+ \in \Up^{W,L'''}$, 
where $L''' = 1+2(L'+L'')$, and therefore $n(W,L''')\asymp W^2L'''\leq  c_1 (k+m) +c_2W^2$.
  
 Now, in the case of a general pattern $S$ with $k$ breakpoints,  
we write $S = S_+ -S_-$,
where $S_+, S_-$ are nonnegative, vanish outside $[0,1]$, and have $k' \le 2k$ breakpoints. 
We also decompose each sum~\eqref{fractal} corresponding to $S_+$ and $S_-$
into a sum over odd indices and a sum over even indices to guarantee disjointness of the  underlying intervals.
In this way, $F$ is represented as a  sum of  the $\Relu$ of four  functions
of the form $(S_i\circ T_i - \hat{S}_i \circ \hat{T}_i)$
each of them belonging to $\Up^{W-2,2(L'+L'')}$  and according to Proposition \ref{composnetsup},
 it follows that $F \in \Up^{W,L}$,
where $L  =4+8(L'+L'')$.
Finally, a parameter count gives
$$
n(W,L) \asymp  W^2L=4W^2+8W^2(L'+L'')\leq  C_1 (k+m) +C_2W^2,
$$ 
 where $C_1$ and $C_2$ are absolute constants and concludes the proof.
\hfill$\Box$

 \begin{remark}
 \label{piecingremark}  
 %The above constructions hold equally well if $S\in \Upsilon^{W-2,ck}$ and $S(0)=S(1)$.
 The  above argument also works 
if the condition $S\in \cS({\bf \xi})\subset  \Sigma_k$ is replaced by 
 $S \in \Upsilon^{ W-4,L}$,  where $S(0)=S(1)$ and  $n( W-4,L) \le Ck$,
with $C$ being an absolute  constant.
 \end{remark}

%%%%%%%%%%%%%%%

%%%%%%%%%%%%%%%%%%%%%%%%%%%%%%%%%%%
%%%%%%%%%%%%%%%%%%%
%%%%%%%%%%%%%%%
%%%%%%%%%%%%%%%%%

\section{$\Relu$ networks are at least as expressive as  Fourier-like sums}
\label{dictionary}

In this section,
we show that $\Relu$ networks can efficiently produce linear combinations of functions from a certain Riesz basis that emulates the trigonometric basis. 
The main point to emphasize here is that the  linear combinations we consider can involve any of these basis functions not just the first consecutive ones.    
Such a linear combination consisting of $n$ basis functions is commonly referred to as an $n$ term approximation from a dictionary (a basis in our case).  
Approximation by such sums is a classic example of nonlinear approximation.

\begin{figure}[ht]
  \centering
\includegraphics[width=.6\textwidth]{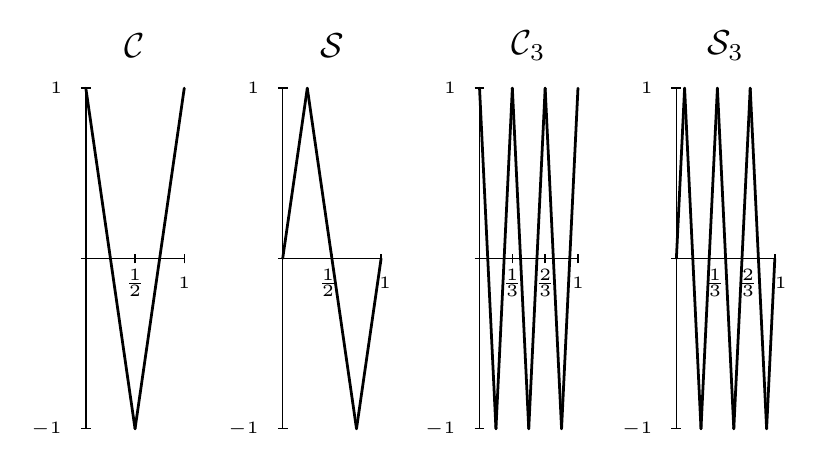}
%\includegraphics[width=.2\textwidth]{figure11.pdf}
%\vspace*{0.5 cm}
\caption{The graphs of ${\cal C}$, ${\cal S}$, ${\cal C}_3$, and ${\cal S}_3$.
}
\label{CS}
\end{figure}

To describe the Riesz basis we have in mind,  we
consider the functions ${\cal C},{\cal S}: [0,1] \to \R$, given by 
$$
{\cal C}(x) :=  \left\{
\begin{matrix}
1- 4x, & x \in [0,1/2),\\
4x - 3, & x \in [1/2,1],
\end{matrix}
\right.
\qquad
{\cal S}(x) :=
 \left\{
\begin{matrix}
4x, & x \in [0,1/4),\\
2-4x, & x \in [1/4,3/4),\\
4x-4, & x \in [3/4,1].
\end{matrix}
\right.
$$
Next, for each $k \ge 1$, we introduce  ${\cal C}_k,{\cal S}_k: [0,1] \to \R$,  defined for any $x \in [0,1]$ by
$$
{\cal C}_k(x) := {\cal C}(kx - \lfloor kx \rfloor),
\qquad
{\cal S}_k(x) := {\cal S}(kx - \lfloor kx \rfloor).
$$
Examples of  representatives of this family of  functions are depicted in  Figure \ref{CS}.
 The system ${\cal F}:=({\cal C}_k,{\cal S}_k)_{k \ge 1}$ is an important example of a family  of 
CPwL functions, since  it  forms
 a Riesz basis for $L_2^0[0,1]$, the set of square integrable functions on $[0,1]$ with zero mean. 
Namely, the following statement holds.
\begin{proposition}
\label{PropRiesz}
The system $({\cal C}_k,{\cal S}_k)_{k \ge 1}$ is a Riesz basis for $L_2^0[0,1]$,
  that is  it spans $L_2^0[0,1]$ and 
 there are absolute constants $c,C>0$ such that,
for any  two   sequences $a,b \in \ell_2(\N)$ of real numbers we have,
\begin{equation}
\label{Frame}
c \sum_{k \ge 1} (a_k^2 + b_k^2)
\le 
\bigg\|
\sum_{k\ge 1} (a_k {\cal C}_k + b_k {\cal S}_k)
\bigg\|_{L_2[0,1]}^2
\le 
C \sum_{k \ge 1} (a_k^2 + b_k^2).
\end{equation}
\end{proposition}
 {\bf Proof:} The proof of this statement is deferred to the appendix.\hfill $\Box$

The following theorem shows how we can produce via $\Relu$ networks $2k$-term linear combinations of elements 
from ${\cal F}$   with a good control on the depth $L$.

\begin{figure}[h]
  \centering
\includegraphics[scale=0.7]{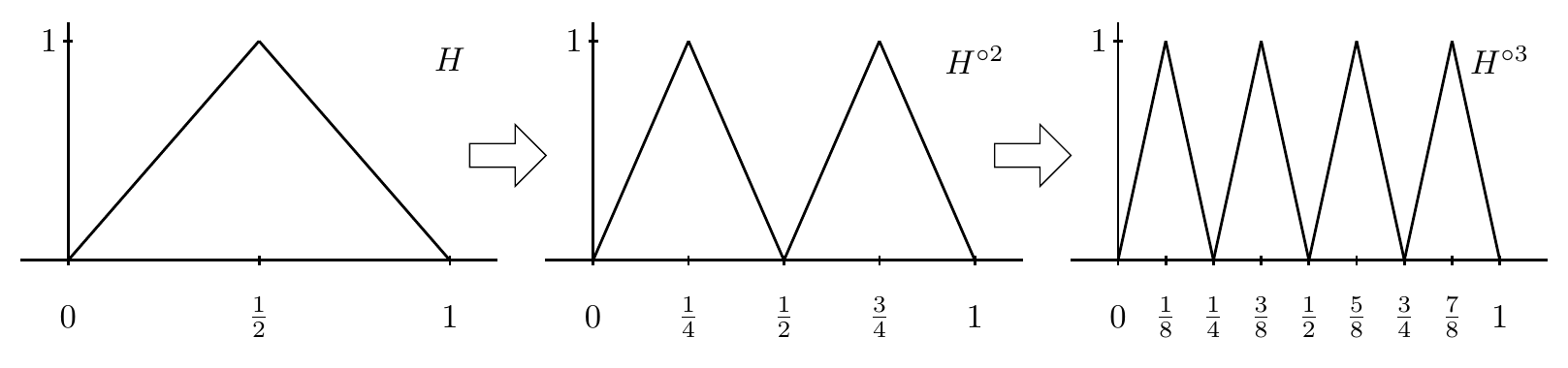}
\caption{The graphs of $H$, $H^{\circ 2}$, and $H^{\circ 3}$.}
\label{Fig16}
\end{figure}

\begin{theorem}
Let $W \ge 6$. For every $k \ge 1$, and set of indices $\Lambda\subset \N$ with $|\Lambda|=k$,  the set
\be
%\label{SetTrigLike}
\nonumber
{\cal F}_\Lambda:=\bigg\{ \sum_{j \in \Lambda}  (a_j \mathcal{C}_j + b_j \mathcal{S}_j ), \;
 a_j,b_j \in \R,  j\in \Lambda, \,\ |\Lambda| = k \bigg\}\subset \Upsilon^{W,L},
\ee
where $\Upsilon^{W,L}$ is produced by a $\Relu$ network
of depth 
$$
L = 2\left\lceil \frac{k}{\lfloor \frac{W-2}{4} \rfloor} \right\rceil(\lceil \log_2(\lambda) \rceil +2),
\qquad \quad\text{with}\quad
\lambda:=\max\{j:\,j\in\Lambda\}.
$$
\end{theorem}
 {\bf Proof:}
With $H$ denoting the hat function from Figure \ref{F:hat-function},
we observe that $H^{\circ m} = H \circ \cdots \circ H$ is a sawtooth function, see Figure \ref{Fig16},
i.e., a CPwL function taking alternatively the values $0$ and $1$ at its breakpoints
 $\ell 2^{-m}$, $\ell = 0,1,\ldots, 2^m$.
Note that  the restriction of the function  $(2^m x - \lfloor 2^m x \rfloor)$ on each interval $[\ell 2^{-m},(l+1) 2^{-m})$   is a linear function
 passing through $\ell 2^{-m}$  with slope $1$. Since ${\cal C}(0)={\cal C}(1)$, one can easily see that 
$$
\mathcal{C}_{2^m}(x)= {\cal C}(2^mx - \lfloor 2^mx \rfloor)= \mathcal{C}(H^{\circ m}(x)).
$$

Since $H$ and $\mathcal{C}=1-2H$ can both be produced by $\Relu$ networks of width $2$ and depth $1$,
it follows from \eqref{Up:pure-compos} that 
$\mathcal{C}_{2^m} \in \Upsilon^{2,m+1}$,   $m=0,1,\dots$.

Next, given an integer  $j$, we find the smallest $m$ with the property $j \le 2^m$.
In view of $\mathcal{C}_j(x) = \mathcal{C}_{2^m}(j 2^{-m} x)$, $j\leq 2^m$, 
we also derive that $\mathcal{C}_{j} \in  \Upsilon^{2,m+1}=\Upsilon^{2,\lceil \log_2j \rceil+1}$.
Likewise, 
because $\mathcal{S}$ can be produced by a $\Relu$ network of width~$2$ and depth~$2$ (by virtue of the identity $\mathcal{S}(x) = \mathcal{C}_2(x/2+3/8)$, $x \in [0,1]$),
we can show that $\mathcal{S}_{j} \in \Upsilon^{2,m+2}=\Upsilon^{2,\lceil \log_2j \rceil+2}$.
Thus, we have established that, 
according to \eqref{Up:full-compos},  for each $j\in \Lambda$,
$$
a_j \mathcal{C}_j + b_j \mathcal{S}_j \in \Up^{4,2\lceil \log_2j \rceil +4}
\subset \Upsilon^{4,2(\lceil \log_2\lambda \rceil +2)}, \quad\text{where}\quad \lambda:=\max\{j:\,j\in\Lambda\}.
$$
Let us denote by $p:=2(\lceil \log_2\lambda \rceil +2)$.
By stacking networks on top of each other,
 a sum of $\lfloor \frac{W-2}{4} \rfloor$ terms $a_j \mathcal{C}_j + b_j \mathcal{S}_j$
belongs to the set $\Upsilon^{4\lfloor \frac{W-2}{4} \rfloor,p}
\subset \Upsilon^{W-2,p}$.
%\snew{\underline{Note:} these stacked networks are disconnected: this is where the number of parameters takes a hit. }
Then, again by \eqref{Up:full-compos},
a sum of $k \le  \lceil k/\lfloor \frac{W-2}{4} \rfloor \rceil \times \lfloor \frac{W-2}{4}\rfloor$ elements $a_j \mathcal{C}_j+b_j \mathcal{S}_j$ belongs to $\Upsilon^{W, \lceil k/\lfloor (W-2)/4 \rfloor\rceil p}$,
as announced.
 \hfill $\Box$

\begin{remark}
Describing the set ${\cal F}_\Lambda$ requires $2k$ parameters,
while the number of parameters $n(W,L)$ for the set
 $\Upsilon^{W,L}$ above has the order of $W^2 L \asymp W k \log_2(\lambda)$.
Ignoring the logarithmic factor,
this is comparable with $2k$ only when the width $W$ is viewed as an absolute constant.

We can take another approach and rather than stacking the networks producing ${\cal S}_j$ and ${\cal C}_j$ on the top of each other,
concatenate them into a special network with width $W=4$. This way we will obtain that
$$
{\cal F}_\Lambda\subset \Upsilon^{4,2k(\lceil \log_2(\lambda) \rceil +2)}.
$$
\end{remark}

%%%%%%%%%%%%%%%%%%%%%%%%%%%%%%%
%%%%%%%%%%%%%%%%%%%
%%%%%%%%%%%%%%%
%%%%%%%%%%%%%%%%%

%%%%%%%%%%%%%%%%
%%%%%%%%%%%%%%%%%

\section{Approximation by (deep) neural networks}
\label{approx}

So far, we have seen in \S\ref{sec:Up}, \S\ref{sec:selfsimilar}, and \S\ref{dictionary}
that $\Relu$ networks can  produce free knot linear splines, self-similar functions, and expansions in 
  Fourier-like Riesz basis of CPwL functions
 using  essentially the same  number of parameters  that are used to describe these sets.
This implies that $\Relu$ networks are at least  as expressive as any of these sets of functions.
In fact, they are  at least  as expressive as the union of these sets,
which intuitively forms a powerful incoherent dictionary. 

   We are more interested in the approximation power of deep neural networks rather than their expressiveness.  Of course, one
 expects these two concepts are closely related.
The remainder of this paper aims at providing    convincing results about the approximation power of $\Relu$ networks  that establishes their superiority
over the existing and more traditional methods of approximation.
We shall do so by concentrating on special $\Relu$ networks $\Up^{W+2,L}$ with a fixed width $W+2$.
 We introduce the notation
\be
%\label{E:upsilon-star}
\nonumber
\Up_m:=\Up^{W+2,m}\subset \Upsilon^{W+2,m},
\qquad \mbox{when} \,\, m\geq 1,
\ee
 and $\Up_0:=\{0\}$, 
and   formally define the approximation  family
$$
\Up:=(\Up_m)_{m\ge 0}.
$$
The number of parameters determining the set $\Up_m$ is $n(W+2,m)\asymp W^2m$, and in going further, we shall refer to them as
{\it roughly}  $W^2m$.  Recall that according to  Proposition~\ref{prnew}, this nonlinear family possesses the following 
favorable properties:
\begin{itemize}
\item Nestedness: $\Up_{m'}\subset \Up_m$  when  $m'\leq m$;
\item Summation property: $\Up_{m'}+\Up_m\subset \Up_{m'+m}$.
\end{itemize}

\subsection{Nonlinear approximation}
\label{nonlinearap}

Let $X$ be any Banach space of functions defined on $[0,1]$.  The typical examples of $X$ are the $L_p[0,1]$ spaces, $1\le p\le \infty$, $C[0,1]$, 
Sobolev and Besov spaces.   Our only stipulation on $X$, at this point,  is that 
it should contain all continuous piecewise linear functions on $[0,1]$.  Given $f\in X$, we define its approximation error when using  deep neural networks to be
\be
%\label{ae}
\nonumber
\sigma_m(f, \Up)_X:=\inf_{S\in\Up_m}\|f-S\|_X, \quad m\ge 0.
\ee
Since  $\Up_0:=\{0\}$, we have  $\sigma_0 (f,\Up)=\|f\|_X$. Given a compact subset $K\subset X$, we define the performance on $K$ to be
\be
%\label{aeclass}
\nonumber
\sigma_m(K, \Up)_X:=\sup_{f\in K} \sigma_m(f,\Up)_X,
\quad m \ge 0.
\ee
In other words, the approximation error on the class $K$ is the worst error.

In a similar way, we define approximation error for other approximation families,
in particular $\sigma_m(f,\Sigma)_X$ and  $\sigma_m(K,\Sigma)_X$ 
when $\Sigma:=(\Sigma_m)_{m\ge0}$ is the family of continuous piecewise linear functions.
We want to understand the decay rate  of $(\sigma_m(f,\Up)_X)_{m \ge  0}$ for individual functions $f$ and of $(\sigma_m(K,\Up)_X)_{m \ge 0}$ for compact classes $K\subset X$ 
and to compare them with the decay rate for other methods of approximation.  

Another  common way to understand the approximation power of  a specific method of approximation such as neural networks   is to characterize
 the following approximation classes.  
 Given $r>0$, the approximation class           
$\cA^r(\Up)_X$, $r>0$, is defined  as the set of all functions $f\in X$ for which
\be
 %\label{ac}
\nonumber
\|f\|_{\cA^r(\Up)_X}:= \sup_{m\ge 0} (m+1)^r\sigma_m(f,\Up)_X,
 \ee
is finite.  
While approximation rates other than $(m+1)^{-r}$ are also interesting, understanding the classes $\cA^r$, $r>0$, 
matches many applications in numerical analysis, statistics, and signal processing. 
The approximation spaces $\cA^r(\Up)_X$ are linear spaces.  
Indeed, if $f,g\in\cA^r(\Up)_X$ and 
$S_m,T_m\in\Up_m$ provide the approximants to $f,g$ satisfying
\be
%\label{ls}
\nonumber
\|f-S_m\|_X\le M(m+1)^{-r}\quad {\rm and} \quad \|g-T_m\|_X\le M' (m+1)^{-r},\quad m\ge 0,
\ee
then $S_m + T_m$ provides an approximant to $f+g$ satisfying
\be
%\label{ls}
\nonumber
\|f+g-(S_m+T_m)\|_X\le (M+M')(m+1)^{-r}\leq 2^r(M+M') (2m+1)^{-r},\quad m\ge 0.
\ee
Since $S_m+T_m$ is in  $\Up_{2m}$ , 
we derive that $f+g\in\cA^r(\Up)_X$.  
We notice in passing that $\|\cdot\|_{\cA^r(\Up)_X}$ is a quasi-norm.

Approximation classes are defined for other methods of approximation in the same way as for neural networks.  
Thus,  given a sequence ${\cal X}:=(X_m)_{m\ge 1}$ of spaces (linear or nonlinear), 
we define $\cA^r({\cal X})_X$ as above 
with $\Up$ replaced by ${\cal X}$.   
The approximation spaces for all classical linear methods of approximation  have 
been characterized for all $r>0$ when $X = L_p[0,1]$ space, $1\le p <\infty$, 
and  $X=C[0,1]$.   
For example, 
%for functions of one variable,
these approximation classes are known for approximation by algebraic polynomials, 
%of order $n$ (degree $n-1$),  
%for approximation 
by trigonometric polynomials, and 
%for approximation 
by piecewise polynomials on an equispaced partition.  
Interestingly enough,  these characterizations do not expose any advantage of one classical linear method over another.
All of these approximation methods have essentially the same approximation classes.  
For example, the approximation classes $\cA^r$ for approximation in $C[0,1]$ by piecewise constants on equispaced partition of $[0,1]$  are the ${\rm Lip} \ r$ spaces when $0<r\le 1$.  
Here, the space Lip $r$ is specified by the condition 
$$|f(x)-f(y)|\le M|x-y|^r$$
and the smallest $M\ge 0$ for which this holds is by definition the semi-norm $|f|_{{\rm Lip} \ r}$.  
The space~$\cA^r$, $0<r<1$, remains the same if we use
trigonometric polynomials of degree $m$.  
The notion of Lipschitz spaces can be extended to $r>1$ and then can be used to characterize approximation spaces
$\cA^r$  when $r>1$.   
We do not go into more detail on  approximation spaces 
for the classical linear spaces but we refer the reader to \cite{DL} for a complete description. 

The situation changes dramatically when using nonlinear methods of approximation.  
There is typically a huge gain in favor of nonlinear approximation in the sense that their approximation classes are much larger than for linear approximation, and so it is easier for a function to have the approximation order $O(m^{-r})$.
 We give just one example,
 important for our discussion of neural networks,  to pinpoint this difference. 
 It is easy to see that any continuous function of bounded variation is in $\cA^1(\Sigma)$.
 Namely, given such a target function $f$ defined on $[0,1]$ and with total variation one, we partition $[0,1]$ into $m$ intervals such that the variation of $f$ on each of these intervals is  $1/m$.   
 Then, the CPwL function which interpolates $f$ at the endpoints of these intervals is in $\Sigma_m$ and approximates $f$ with error at most~$1/m$.   
 Notice that such functions of bounded variation are far from being in
${\rm Lip} \ 1$ because they can change values quite abruptly.   
This illustrates the central theme of nonlinear approximation that their approximation spaces are 
much larger than their linear counterparts.  We refer the reader to \cite{DNL} for an overview of nonlinear approximation.

%%%%%%%%%%%%%%%%%%%
%%%%%%%%%%%%%%%
%%%%%%%%%%%%%%%%%

\subsection{Approximation of classical smoothness spaces}
\label{sec:classical} 

Let us start this section by revisiting Theorem \ref{T:main}, which states that 
$$
\Sigma_m\subset   \Upsilon^{W,\frac{C}{W^2}m}, \quad m \geq q(W-2),
$$
and 
$$
\Sigma_m\subset \Up^{W,2}, \quad 1\leq m< q(W-2),
$$
where $q=2$ when $2\leq W<7$ and  otherwise $q=\lfloor\frac{W-2}{6}\rfloor$. This follows from the simple observation that
 $W^2L\asymp n(W,L)\leq Cm$, when $m\geq q(W-2)$. In addition, for any $m$ we can embed  
$\Upsilon^{W,m}\subset\Up^{W+2,m}= \Up_m$ by  adding a source and collation channel. 
Hence, in the view of the new notation,  Theorem \ref{T:main} can be restated the following way.
\begin{theorem}
\label{T:mainrestated}
For $m\geq q(W-2)$, we have 
$$
\Sigma_m\subset \Up_{\gamma m}, \quad \text{where}  \quad \gamma=\gamma(W)=\frac{C}{W^2},
$$
and thus for any $f\in C[0,1]$
$$
\sigma_{\gamma m}(f,\Up)_{C[0,1]}\leq \sigma_{m}(f,\Sigma)_{C[0,1]}.
$$
For $1\leq m<q(W-2)$, 
$$
\Sigma_m\subset \Up_{2},
$$
 and thus  for any $f\in C[0,1]$ we have $\sigma_{2}(f,\Up)_{C[0,1]}\leq \sigma_{m}(f,\Sigma)_{C[0,1]}$.
\end{theorem}
Therefore, all approximation results that involve the error of best approximation $\sigma_{m}(f,\Sigma)_{C[0,1]}$ 
by the family $\Sigma$ of free knot linear splines 
will hold for the error of best approximation $\sigma_{\gamma m}(f,\Up)_{C[0,1]}$ by the family $\Up$.

While we do not expect improvement in the approximation power of classical smoothness classes when using neural networks, there is a little twist here that was exposed in the work of Yarotsky~\cite{Y2}.  
 He  proved that for $W=5$,
$$
\sup_{f\in {\rm Lip} \ 1}\inf_{S\in \Upsilon^{W,m}}\|f-S\|_{C[0,1]}\le C\frac{|f|_{{\rm Lip} \ 1}}{m\ln m}.
$$
Since $\Upsilon^{W,m}\subset\Up_m$ (by just adding a source and collation channel), his result can be restated 
using our  notation as
the following result for approximating functions in ${\rm Lip} \ 1$ by $\Relu$ networks
\be
\label{ex3}
\sigma_m(f,\Up)_{C[0,1]} \le  C(W) \frac{|f|_{{\rm Lip} \ 1}}{m\ln m},\quad m\ge 2,
\ee
 in the particular case $W=5$. Note that the number of parameters  describing
$\Up_m$ is roughly $W^2m$ , and 
 the surprise in  \iref{ex3} is the favorable  appearance of the logarithm.
Indeed, for all other  standard methods of linear or nonlinear approximation depending on  $Cm$ parameters, including $\Sigma$, there is a function $f\in{\rm Lip} \ 1$ which cannot be approximated with accuracy better than $c/m$, $m\ge 1$.

%%%%%%%%%%%%%%%%%%%
%%%%%%%%%%%%%%%
%%%%%%%%%%%%%%%%%

\subsubsection{The space Lip $\alpha$}
\label{ss:Lip}

 Yarotsky's theorem can be generalized in many ways.
We begin by discussing the Lip $\alpha$ spaces.  
For this, we isolate a simple remark about the Kolmogorov entropy of the unit ball  of ${\rm Lip} \ \alpha$.  
Let $K_\alpha$ be the set of functions with $|f|_{{\rm Lip}\ \alpha}\le 1$ vanishing at the endpoints~$0$ and $1$.
 
 \begin{lemma}
 \label{entropylemma}
 For each $0<\alpha\le 1$ and for each integer $k\ge 2$, there  are   at most $3^k$ patterns 
$S_1,\dots, S_{3^k}$ from $\cS(\xi)$, $\xi=(0,\frac{1}{k},\dots,\frac{k-1}{k},1)$,  such that whenever $g\in K_\alpha$,  there is a $j\in\{1,\dots,3^k\}$ with
 \be
 \label{entropy1}
 \|g-S_j\|_{C[0,1]}\le   2h^\alpha, \quad h:=\frac{1}{k}.
 \ee
 In other words, the set $K_\alpha$ can be covered by $3^k$ balls in $C[0,1]$ of radius $2k^{-\alpha}$ with centers from~$\cS(\xi)$.
 \end{lemma}
 \noindent
 {\bf Proof:}    
 We consider the following set $\cP$ of patterns from $\cS(\xi)$.  
 For $T$ to be in $\cP$, we require that $T(\xi_j)=m_j h^\alpha$,  with  $m_0,\dots, m_k$   integers satisfying the conditions
\be
\label{Condm}
m_0=m_k=0, \quad |m_j-m_{j-1}|\le 1, \quad j=1,\dots,k.
\ee
There are at most $3^k$ such patterns, i.e.,  $\#(\cP)\le 3^k$.  

For the proof of our claim, 
given $g\in K_\alpha$,
we first notice that  $|g(\xi_j)-g(\xi_{j-1})|\leq h^\alpha$, $j=1,\dots,k$.  
We then approximate $g$ by    the CPwL function $S\in\cS(\xi)$, 
%{\rnew defined as follows.
 where the values  
 $S(\xi_j)$  are  of the form $ \beta_j h^\alpha$, $ \beta_j \in \mathbb{Z}$,
 and are chosen so that $S(\xi_j)=\beta_jh^\alpha$ is the closest to $g(\xi_j)$, 
$j=1,\ldots,k$. Note that this gives   $\beta_0=\beta_k=0$ since $g(\xi_0)=0=g(\xi_k)$ and 
\be
\label{ll}
\left|  S(\xi_{j})-g(\xi_j) \right|\leq  h^\alpha/2.
\ee When 
 assigning the  values $S(\xi_j)$, starting with $S(\xi_0)=0$ and moving from left  to  right, if it happens that 
there are two possible choices for $\beta_j$ (which happens if $g(\xi_j)\pm h^\alpha/2$ is an integer multiple of 
$h^\alpha$), we select the $\beta_j$ that is closest to the already determined  $\beta_{j-1}$.
Since
\begin{eqnarray*}
\left| \beta_j  - \beta_{j-1}  \right| h^\alpha
 & = &  \left| S(\xi_j) - S(\xi_{j-1})  \right| \\
\nonumber
 &\leq & \left|  S(\xi_{j})-g(\xi_j) \right|+
 \left| g(\xi_j) - g(\xi_{j-1}) \right|
+\left| g(\xi_{j-1}) - S(\xi_{j-1}) \right|\\
& \le & h^\alpha/2 + h^\alpha +  h^\alpha/2 = 2 h^\alpha,
\end{eqnarray*}
we have  $|\beta_j-\beta_{j-1}|\le 2$. But the case of equality is not possible
since it would mean that at step $j$ we have not selected $\beta_j$ to be the closest to $\beta_{j-1}$. 
Therefore $|\beta_j-\beta_{j-1}|\le 1$, and thus 
 \eqref{Condm} holds,
i.e., the constructed approximant $S$ is a pattern from $\cP$.
 Finally, 
we notice that any pattern from $\cP$ has slopes with absolute value at most  $h^{\alpha -1}$.
Hence, for any  $x \in [0,1]$, picking the point $\xi_j$  the closest to $x$,
we have
$$
 |g(x)-S(x)| \le
 |g(x)-g(\xi_j)| + |g(\xi_j)-S(\xi_j)| + |S(\xi_j)-S(x)|
 \le (h/2)^\alpha + h^\alpha /2 + h^{\alpha - 1} (h/2) \le 2 h^\alpha,
$$
 where we used \eref{ll} and the fact that $|x-\xi_j|\le h/2$.  
Taking the maximum over $x \in [0,1]$ establishes \eref{entropy1}
and concludes the proof.
 \hfill $\Box$

 The following theorem generalizes \eref{ex3} to Lip $\alpha$ spaces.
\begin{theorem}
\label{Liptheorem}  
 Let $W\geq 8$. If $X=C[0,1]$ and  $f\in  {\rm Lip  }\ \alpha $,   $0<\alpha\le 1$, then 
\be
\label{t21}
\sigma_m(f, \Up)_X \le   C(W) \frac{|f|_{{\rm Lip}\ \alpha }} {( m  \ln m)^{\alpha}}, \quad m \ge 2.
\ee
\end{theorem}

\noindent
{\bf Proof:}   
Without loss of generality, we can assume that $|f|_{{\rm Lip} \ \alpha } =1$. 
Fixing $f $ and $m$,  we first choose $T$ as the piecewise linear function
which interpolates $f$ at the equally spaced points $x_0,\dots,x_m$, where $x_i:=i/m$, $i=0,\dots,m$. 
 %We next check the Lip $\alpha$ norm of $T$ on the interval  $I_j:=[x_j,x_{j+1}]$. 
Since $f$ and $T$ agree at the endpoints of the interval $J_i:=[x_i,x_{i+1}]$, 
the slope of $T$ on $J_i$ has absolute value at most  $m^{1-\alpha}$.  
Therefore,
$$
|T(x)-T(y)|  \le   m^{1-\alpha}   |x-y| \le  |x-y|^\alpha,\quad x,y\in J_i,
$$
and hence $T$ is also in Lip $\alpha$ with semi-norm at most one on each of these intervals.

We now  define  $g:=f-T$
and write $g = \sum_{i=1}^m g \chi_{J_i}$.
Each $g_i := g \chi_{J_i}$ is a function in Lip $\alpha$  with $|g_i|_{{\rm Lip} \ \alpha}\le 2$.
Let $k$ be the largest  integer such that 
$3^k k\le m$ and let $\cP=\{S_1,\dots,S_{3^k}\}$ be the set of the $3^k$ patterns given by Lemma \ref{entropylemma}. 
Applying this lemma to each of the functions  $\bar g_i:[0,1]\rightarrow \R$, defined by 
$\bar g_i(x):= 2^{-1}m^\alpha g_i((x+i)/m) \in K_\alpha$,  we find a pattern $S_{j_i}\in \cP$, 
$S_{j_i}:[0,1]\rightarrow \R$, such that 
$$
\|\bar g_i-S_{j_i}\|_{C[0,1]}\leq 2k^{-\alpha}.
$$
Shifting back to the interval $J_i$ provides a function $S_{j_i} \in \cP$
such that 
$$
|g_i(x)- 2 m^{-\alpha} S_{j_i}(m(x-x_i))| \leq  4{(km)^{-\alpha}}, \quad x \in J_i,
$$
and therefore  the function $\hat T$ given by
\be
\label{patternsum}
\hat T(x):=T(x)+ 2m^{-\alpha} \sum_{i=1}^m S_{j_i}(m(x-x_i)) 
\ee
approximates $f$ to accuracy  $4(km)^{-\alpha}$
in the uniform norm.

Since there are $3^k < m$ patterns, 
some of them must be repeated in the sum \eqref{patternsum}.
For each $j = 1,\ldots,3^k$,
we consider the (possibly empty) set of indices 
$\Lambda_j = \{ i \in \{ 1,\ldots,m \}: j_i = j\}$.
We have
\be
%\label{defineTi}
\nonumber
\hat T=T+\sum_{j=1}^{3^k} T_j,
\qquad \mbox{where} \quad
T_j:= 2m^{-\alpha}\sum_{i \in \Lambda_j}  S_{j}(m(x-x_i)).
\ee
Since $T\in \Sigma_m$,   Theorem \ref{T:main} says that 
 $T$ belongs to $\Up^{W,L_0}$ with  either $W^2 L_0\asymp n(W,L_0) \le C' m$ or $L_0=2$.
According to  Lemma \ref{fractallemma},
each function $T_j$ is in   $\Up^{W,L_j}$ with  either $W^2 L_j\asymp n(W,L_j) \le C_1(k+ m_j)+C_2W^2$ or $L_j=2$,
 where $m_j:=|\Lambda_j|$.
Therefore, in view of \eqref{Up:full-compos},
we derive that $\hat T $ belongs to $\Up^{W,L}$ with $L = L_0 + \sum_{j=1}^{3^k} L_j$,  and
$$
 L =  L_0 +  \sum_{j=1}^{3^k} L_j
\le \frac{1}{W^2}\left(C' m + C_1 3^k k + C_1 \sum_{j=1}^{3^k} m_j \right)+C_33^k
\le \left (\frac{\tilde C_1}{W^2}+\tilde C_2\right)m=c(W)m,
$$
where we have used the facts that $3^k k \le m$ and $\sum_{j=1}^{3^k} m_j = m$.
This shows that $\hat T \in \Up_{c(W)m}$ and in turn that
$$
\sigma_{c(W)m}(f,\Up)_{C[0,1]} 
\le \|f - \hat T \|_{C[0,1]}
\le \frac{4}{(km)^{\alpha}}
\le \frac{\wt{C}}{(m \ln m)^{\alpha}}, 
$$
where   in the last inequality we have used that  $k \ge c\ln m$  since 
$3^{k+1} (k+1) > m$.
 Up to the change of $m$ in $c(W)m$,
this is the result announced in \eqref{t21}.
\hfill $\Box$

%%%%%%%%%%%%%%%%%%%
%%%%%%%%%%%%%%%
%%%%%%%%%%%%%%%%%

\subsubsection{Other classical smoothness spaces}
\label{ss:interpolation}
We can also exhibit a certain logarithmic improvement in the approximation rate for functions in other smoothness classes.  
Since we do not wish to delve too deeply into
the theory of smoothness spaces in the present paper, we illustrate this with just one example. 
\begin{theorem}
\label{Wptheorem}  
 Let $W\geq 8$. If $X=C[0,1]$ and $f \in C[0,1]$ satisfies $f'\in L_p[0,1]$, $1\le p\le \infty$, then 
\be
\label{Wp}
\sigma_m(f,\Up)_X \le  C(W) \frac{\|f'\|_{L_p}}{m ( \ln m)^{1-1/p}},\quad m\ge 2, 
\ee
where  $C(W)$ depends also on  $p$ (when $p$ is close to one).
\end{theorem}

\noindent
{\bf Proof:}  
When $p=\infty$, \eref{Wp} follows from Theorem \ref{Liptheorem} since $f'\in L_\infty$ is equivalent to $f\in {\rm Lip} \ 1$ and $|f|_{{\rm Lip} \ 1}=\|f'\|_{L_\infty}$.  The case $p=1$ follows from
\be
\label{p1}
\sigma_{\gamma m}(f,\Up)_X\le \sigma_m(f,\Sigma)_X\le \|f'\|_{L_1} m^{-1},\quad m\ge 1.
\ee
 Here, the first inequality  follows from Theorem \ref{T:mainrestated} 
 and the second inequality is a consequence of an estimate (already mentioned) for CPwL approximation of continuous functions of bounded variation,
 which applies to $f$ since $f'\in L_1$. 
 Now, given $1<p<\infty$ and $f \in C[0,1]$ with $f'\in L_p$, for any $t>0$, we can write 
\be
%\label{interpolation}
\nonumber
f=f_0+f_1,
\ee
where  
\be
%\label{interpolation1}
\nonumber
\max\{\|f_1'\|_{L_1},t \|f_0'\|_{L_\infty} \}\le \|f_1'\|_{L_1} + t \|f_0'\|_{L_\infty}\le C\|f'\|_{L_p}t^{1-1/p},
\ee
and $C$  is a constant depending on $p$ when $p$ is close to $1$.
This is a well-known result in interpolation of operators (see \cite{DSobolev}).
We take $t:=(\ln m)^{-1}$ and find 
\begin{eqnarray}
%\label{sigmaf}
\nonumber
\sigma_{2\gamma m}(f,\Up)_X&\le&  \sigma_{\gamma m}(f_0,\Up)_X+\sigma_{\gamma m}(f_1,\Up)_X \nonumber \\
&\le &  C(W) \{ \|f_0'\|_{L_\infty} (m\ln m)^{-1} + \|f_1'\|_{L_1}m^{-1} \}\nonumber \\
&\le & C(W)\|f'\|_{L_p} \{ (m\ln m)^{-1} t^{-1/p} + m^{-1}t^{1-1/p}\}    \nonumber \\
&\le &  C(W)\|f'\|_{L_p} m^{-1}(\ln m)^{-1+1/p},
\nonumber
\end{eqnarray}
where we used the summation property for the elements of the family $\Up$.
The second inequality followed from Theorem \ref{Liptheorem} with $\alpha=1$ and from the estimate \eref{p1}.\hfill $\Box$

%%%%%%%%%%%%%%%%%%%
%%%%%%%%%%%%%%%
%%%%%%%%%%%%%%%%%

\subsection{The power of depth}
\label{depth}

 The previous subsection showed that functions taken from classical smoothness spaces typically enjoy some mild improvement in approximation efficiency when using $\Relu$ networks rather than more classical methods of approximation.  However,
 this modest gain does not give any convincing reason for the success of deep networks, at least from the viewpoint of their approximation properties.  
 In this subsection, we highlight several classes of functions whose approximation rates by neural networks far exceed their approximation rates by free knots linear splines  or any other standard approximation family.  
 Our constructions are based on variants of the following simple observation.
 
 \begin{proposition}
 \label{sumlemma} 
For functions $f_k\in \Up_k$ satisfying $\|f_k\|_{C[0,1]}=1$ for all $k\ge 1$ 
 and for a sequence $(\beta_k)_{k\ge 1}$ in $\ell_1(\N)$, the function
 \be
 %\label{sum1}
\nonumber
 F:=\sum_{k\ge 1}\beta_k f_k
 \ee
 has approximation error satisfying 
 \be
 %\label{errorF}
\nonumber
 \sigma_{m^2}(F,\Up)_{C[0,1]}\le \sum_{k>m}|\beta_k|, \quad m\geq 1.
 \ee
 \end{proposition}
 
 \noindent
 {\bf Proof}.  The function $S_m:=\sum_{k=1}^m  \beta_k f_k$ belongs to $\Up_{ m^2}$,  thanks to the summation  and inclusion properties for $\Up$.   
 A triangle inequality gives
 $$
 \|F-S_m\|_{C[0,1]}\le  \sum_{k>m}|\beta_k|,
 $$
 and the statement follows immediately.
\hfill $\Box$.  

\begin{remark}
\label{composeremark}
When the functions $f_k$ are related to one another, 
the proposition can be improved by replacing $m^2$ with a smaller quantity.  
For example,
if $f_k=\phi^{\circ k}$ for a fixed function $\phi$ in 
$\Upsilon^{{\rm w},\ell}$, with width $2\leq {\rm w} \le W-2$,  and fixed depth $\ell$, 
 then Proposition {\rm \ref{cor1}} reveals that $m^2$ can be changed to  $\ell m$.
\end{remark}

We now present some classes of such functions $F$ that are well approximated by $\Relu$ networks.   
For the most part, these functions
cannot be well approximated by  standard approximation families.

%%%%%%%%%%%%%%%%%%%
%%%%%%%%%%%%%%%
%%%%%%%%%%%%%%%%%
\subsubsection{The Takagi class of functions}
\label{Takagiclass}

For our first set of examples, let us recall that functions of the form
\be
\label{dynamical}
F= \sum_{k \ge 1} t^k g(\psi^{\circ k}),\quad |t|<1,
\ee
with $\psi:[0,1]\to[0,1]$ and $g:[0,1] \to \R$, provide primary examples of self similar functions and dynamical systems~\cite{YH}. 
If $g\in \Upsilon^{W_1,\ell}$ and $\psi\in \Upsilon^{W_2,\ell}$, with 
$W_1+W_2 = W$,
Proposition \ref{cor2} implies that the partial sum $S_m:=\sum_{k=1}^m t^k g(\psi^{\circ k})$
belongs to  $\Up^{W,\ell(m+1)}\subset \Up_{\ell(m+1)}$. 
Therefore, in this case, 
the function $F$ defined via \eref{dynamical} is approximated 
by the partial sum $S_m$ with exponential accuracy by $\Relu$ networks,  that is
$$
\sigma_{\ell(m+1)}(F,\Up)_{C[0,1]}\leq Ct^{m+1}, \quad |t|<1.
$$

\begin{figure}[h]
  \centering
\includegraphics[scale=1.0]{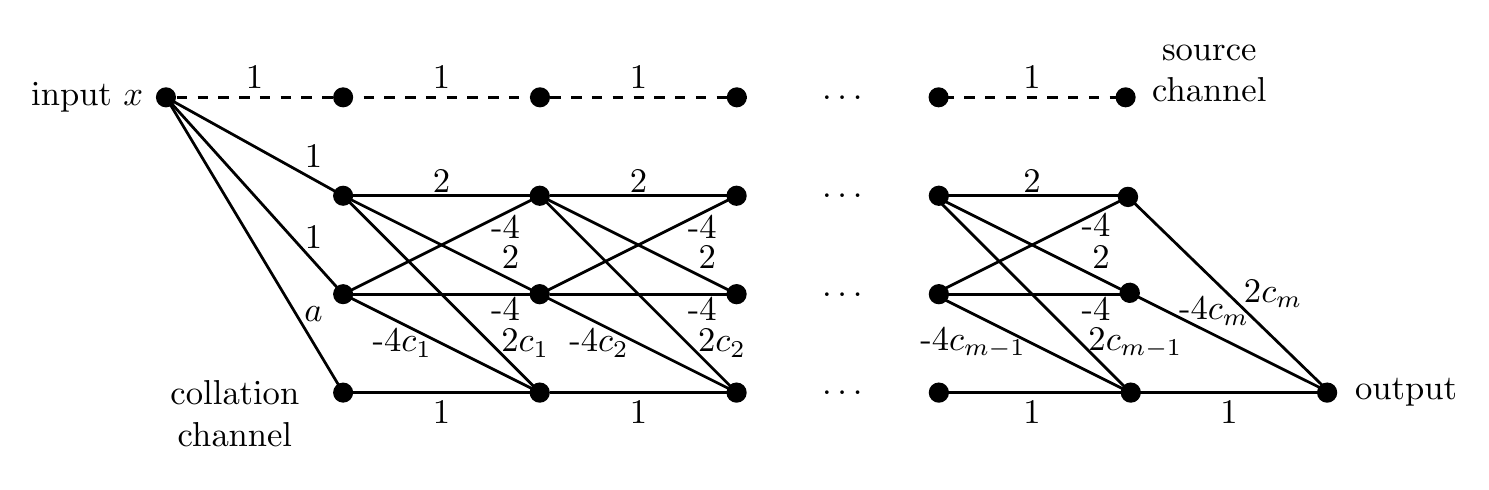}
\caption{The computation graph associated to 
 the approximation of the Takagi class.}
\label{Fig12}
\end{figure}

 Now,  we consider a special class of functions. For this purpose,  
we recall that the hat function  $H\in \Upsilon^{2,1}$ and its $k$-fold composition
$H^{\circ k}:= H\circ H\circ \cdots\circ H$,  according to the composition property \eref{Up:pure-compos},
belongs to $\Up^{2,k}$. On the other hand,  the same function
$H^{\circ k}$ is  in $\Sigma_n$ only if $n$ is exponential in $k$.
For an absolutely summable sequence $(c_k)_{k\ge 1}$ of real numbers, 
we consider continuous functions  $F$ of the form 
\be
%\label{hatex}
\nonumber
F:=  \sum_{k\ge 1} c_k H^{\circ k},
\ee
approximations to which are produced by the special $\Relu$ networks shown in Figure~\ref{Fig12}.
%Each such function is in $C[0,1]$.  
The collection of all such functions is called the Takagi class. 
It contains a number of interesting and important examples.  
A good source of information on the Takagi class is \cite{AK},
 from which the two examples below are taken. 
 
For the first example, we take  $c_k:= 2^{-k}$, which gives the Takagi  function
\be
%\label{Tfunction}
\nonumber
T:=\sum_{k\ge 1} 2^{-k}H^{\circ k}.
\ee
From Remark \ref{composeremark}, we have
\be
%\label{Terror}
\nonumber
\sigma_{m}(T,\Up)_X\le 2^{-m},\quad m\ge 1,
\ee
and so  theoretically $T$ can be approximated with exponential accuracy by $\Relu$ networks with  roughly $W^2m$ parameters.  
In practice, see Figure \ref{Fig12}, we can approximate it using $m$ parameters.
However, $T$ is nowhere differentiable and so it has  very little smoothness in the classical sense.   
This means that all of the traditional methods of approximation will fail  miserably to approximate it.  
Note that the function $T$ has self similarity, in that it satisfies a simple refinement equation. 

 %%%%%%%%%%%%%%%
Other examples take a highly lacunary sequence of coefficients  and  thereby  construct functions  in the Takagi  class 
that do  not  satisfy  a  Lipschitz  condition  of  any  order  and yet they can be approximated to exponential accuracy by $\Up$.
Many functions from the Takagi class are fractals, in the sense that the
Hausdorff dimension of their graph is strictly greater than one.

We do not go into the Takagi class more deeply but refer the reader to  \cite{AK,H}  where the properties and 
applications of the Takagi functions are given as well as numerous examples of  similar constructions.  
The main point to draw from these examples is that the approximation classes $\cA^r$ for $r$ large contain many functions which are not smooth in any classical sense.

%%%%%%%%%%%%%%%%%%%
%%%%%%%%%%%%%%%
%%%%%%%%%%%%%%%%%

\subsubsection{Analytic functions} 
 Another example in the Takagi class is  the function 
\be
\label{power1}
x(1-x)=\sum_{k\ge 1} 4^{-k} H^{\circ k}.
\ee
This formula is used as a starting point to show that analytic functions are well approximated by deep neural networks (see \cite{Y1,LS,E}), as we briefly discuss below.

It follows from \eref{power1} that the function $x^2$ is approximated with exponential accuracy by $\Relu$ networks.  
From this, one derives that all power functions $x^k$ also are approximated with exponential accuracy.     
Then, using the summation property, 
one concludes that analytic functions and functions in Sobolev spaces are approximated
with the same accuracy as their approximation by algebraic polynomials.  
Similarly, we can approximate functions on $[0,1]$ from their power series representation.
The point we emphasize here is the flexibility of $\Relu$ networks,
 in that they approximate well functions with little classical smoothness but
  retain the property of approximating classically smooth functions with the same accuracy as other methods of approximation.

 %%%%%%%%%%%%%%%%%%%
%%%%%%%%%%%%%%%
%%%%%%%%%%%%%%%%%

%%%%%%%%%%%%%%%%%%%%%%
 
\section{Neural network approximation as manifold approximation}
\label{sec:manifold}

Up to this point, we reflected the expressive power and the corresponding approximation power  of deep ReLU networks.  
In other words, we wondered how well the best approximation from $\Up_m$
to a target function performs.
An important practical issue is the construction of reasonable methods of approximation that yield near-best approximations to any given target function  $f\in X$ with e.g. $X=C[0,1]$.    
 
 To discuss this problem, we need to formulate what would be considered a reasonable approximation procedure.   
 The set $\Up_{m}$ is described by  roughly $W^2m$  parameters,
 which are identified by a point in $\R^m$.  
 We  let $M=M_m$ be the mapping that sends $z\in \R^m$ to the function $M(z) $ generated by the neural network with the chosen parameters $z$.  
We view the collection $\cM = \cM_m$ of all $M(z)$, $z\in\R^m$, 
 as an $m$-dimensional  manifold.  
 In this context, we also view any approximation method as providing a mapping $a=a_m:X\to \R^m$
 which, for a given $f\in X$, selects the parameters of the network
 used to approximate $f$.  
 The approximation to $f$ is then
 \be
% \label{approxf}
\nonumber
 A_m(f)=M_m(a_m(f)),\quad n\ge 0.
 \ee
 A fundamental question for both theory and numerical practice is what conditions to impose on $a_m$ and $M_m$ so that the resulting scheme 
$A_m$ is reasonable.  
 In keeping with the notion of numerical stability,
 we could  require that each of these mappings is a Lip $1$ with a fixed constant $\Gamma$ independent of $m$.  
This means that there is a norm $\|\cdot\|$ on $\R^m$ (typically an $\ell_p$ norm) such that,
 for any $f_1,f_2\in X$, 
 \be
 %\label{stable1}
\nonumber
 \|a_m(f_1)-a_m(f_2)\|\le \Gamma \|f_1-f_2\|_X.
 \ee
 The stability of $M_m$ means that, for any $z_1,z_2\in \R^m$,
 \be
 %\label{stable2}
\nonumber
 \|M_m(z_1)-M_m(z_2)\|_X\le \Gamma  \|z_1-z_2\|.
 \ee

 One can lessen the demand on numerical stability to requiring only that the mappings $a_m$ and $M_m$ are continuous, not necessarily Lipschitz. 
 This weaker assumption was used
 in the definition of manifold widths \cite{DHM}.  This manifold width of a compact set $K\subset X$ is defined as 
 \be
 %\label{stablewidth}
\nonumber
 \delta_m(K):=\inf_{(a,M)}\sup_{f\in K}\|f-M(a(f))\|_X,
 \ee
where the infimum is taken over all continuous maps.  
It is shown in \cite{DKLT} that this milder requirement still puts a restriction on how well sets characterized by classical smoothness can be approximated.  
For example, if $K$ is the unit ball of Lip $\alpha$,
then  $\delta_m(K)\ge Cm^{-\alpha}$.
Therefore, 
the logarithmic improvement featured in \S\ref{sec:classical} cannot be obtained with continuous selection of parameters.  
This lack of continuity for some approximation schemes was also recognized in \cite{KKV}.  
This may be a crucial point in the framing of recent results on the instability of certain methods for constructing deep network approximations to target
functions from data via optimization methods (such as least squares or constrained least squares methods).

%%%%%%%%%%%%%%%%%% 
%%%%%%%%%%%%%%%%%% 
%%              %% 
%% BIBLIOGRAPHY %% 
%%              %% 
%%%%%%%%%%%%%%%%%% 
%%%%%%%%%%%%%%%%%% 

\section{Appendix }

\subsection{ The matrices of Lemma \ref{core}}
\label{ss1}

 In order to explicitly write the affine transforms $A^{(1)}$ and $A^{(2)}$ that determine the $\Relu$ net, we describe here  
one of the possible ways to partition the set of indices $\Lambda$ so that the constant sign and separation properties are satisfied. To do this, we   first consider  $\Lambda_+$ and only  the 
main breakpoints $\xi_{j}$ with indices $j$ for which  $j\!\!\mod 3=\ell$. We collect  into the set 
$\Lambda_i^{\ell,+}$ all indices $k\in \Lambda_+$ that correspond to   the $i$-th hat function $H_{i,j}$
associated to a principal breakpoint $\xi_{j}$ with the above mentioned property. Recall that there are 
$q$ hat functions $H_{i,j}$ associated to each principal breakpoint $\xi_j$.
We do this for every $\ell=0,1,2$, and $\Lambda_-$, and  we get the partition
\begin{eqnarray}
\nonumber
\Lambda_i^{\ell,+}:=\{s: \, s\in \Lambda_+\,\,\text{and}\,\,\phi_s=H_{i,j}\,\,\text{with}\,\,j \!\!\!\mod 3=\ell\},\\
\nonumber
\Lambda_i^{\ell,-}:=\{s: \, k\in \Lambda_-\,\,\text{and}\,\,\phi_s=H_{i,j}\,\,\text{with}\,\,j \!\!\!\mod 3=\ell\},
\end{eqnarray}
where  $ \ell=0,1,2$, $i=1,\ldots,q$. The matrices that determine the special $\Relu$ network are
$$
M^{(1)}=\begin{bmatrix}1&1&\ldots&1&0\end{bmatrix}^T, \quad
b^{(1)}=\begin{bmatrix}0&\xi_1&\ldots&\xi_{W-2}&0\end{bmatrix}^T,
$$
$$
M^{(2)}=\begin{bmatrix}1&0&\ldots&0&0 \\
                                    m_{2,1}^{(2)}&m_{2,2}^{(2)}&\ldots&m_{2,W-1}^{(2)}&0 \\
\ldots&\ldots&\\
m_{W-1,1}^{(2)}&m_{W-1,2}^{(2)}&\ldots&m_{W-1,W-1}^{(2)}&0 \\
0&0&\ldots&0&1
\end{bmatrix}, \quad 
b^{(2)}=\begin{bmatrix}0\\b_2^{(2)}\\\ldots\\b_{W-2}^{(2)}\\0\end{bmatrix}
$$
$$
M^{(3)}=\begin{bmatrix}0&\varepsilon^{(3)}_1&\ldots&\varepsilon^{(3)}_{W-2}&1\end{bmatrix},\quad b^{(3)}=0,
$$
where $\varepsilon^{(3)}_k=1$ if $\Lambda_k\subset \Lambda_+$, $\varepsilon^{(k)}_k=-1$ 
if $\Lambda_k\subset \Lambda_-$, and $\varepsilon^{(3)}_k=0$ if $\Lambda_k=\emptyset$. 
Next, we demonstrate how to 
find the entrances of one row in $M^{(2)}$. The rest of the rows are computed likewise. The index 
$k=1,\ldots,W-2$ in \eref{decomp} corresponds to a different labeling of the index set 
$$
\{(i,\ell,+), (i,\ell,-),\,i=1,\ldots,q, \, \ell=0,1,2\},
$$
of the particular partition we work with here. We take the index $(1,1,+)$ and compute the corresponding $\tilde T$,
$$
\tilde T:=T_{(1,1+)}=\sum_{s\in \Lambda_1^{1,+}}c_s\phi_s=[\tilde S]_+,
$$
see Figure \ref{Sgraph},
%%%%%%%%%%%%%%%%%%
%%%%%%%%%%%%%%%%%%%
%%%%%%%%%%%%%%%
where $\tilde S$ is a CPwL function with breakpoints the principal breakpoints $\xi_1,\ldots,\xi_{W-2}$, with the property 
$$
\tilde S(\xi_{4s+1})=c_{4s+1}, \quad \tilde S(x_{(4s+1)q-1})=
\tilde S(x_{(4s+1)q+1})=0, \quad s=0,\ldots,\left\lfloor \frac{W-2}{4}\right\rfloor.
$$
Then the  entries in the second row in $M^{(2)}$ and $b^{(2)}$ are the coefficients  from the representation, 
$$
\tilde S(x)=m_{2,1}^{(2)}x+\sum_{j=2}^{W-2}m_{2,j}^{(2)}(x-\xi_j)_++b_2^{(2)}.
$$

%%%%%%%%%%%%%%%%%%%%%%%%%%%%%
%%%%%%%%%%%%%%%%%%%%%%%%%%%%
%%%%%%%%%%%%%%%%%%%%%%%%%%%%%%

\subsection{Theorem \ref{T:main}, Case $4\leq W\leq 7$}
In this case we have to show that 
for every $n\geq 1$ the set $\Sigma_n$ of free knot linear splines with $n$ breakpoints 
is contained in the set $\Upsilon^{W,L}$ of functions produced by width-$W$ and depth-$L$ $\Relu$ networks where
\be
\nonumber 
 L=
  \begin{cases}
   2\left \lceil \frac{n}{2(W-2)}\right\rceil, \quad 
 & n\geq 2(W-2), \\
    2, & n<2(W-2),
  \end{cases}
\ee
and whose number of parameters  
\be
\nonumber 
 n(W,L)\leq
  \begin{cases}
   Cn, \quad 
 & n\geq 2(W-2), \\
    W^2+4W+1, & n< 2(W-2),
  \end{cases}
\ee
where $C$ is an absolute constant.
We start with the case   $W-2=2$.
 Given $n\geq 4$, we choose $L:=\lceil \frac{n}{4}\rceil$. If $n<4L$, we add artificial breakpoints so that we represent 
$T\in \Sigma_n\subset \Sigma_{4L}$  as
$$
T(x)=ax+b+\sum_{j=1}^{4L}c_j(x-\xi_j)_+=ax+b+\sum_{j=1}^{2L}S_j, \quad 
S_j:=c_{2j-1}(x-\xi_{2j-1})_++c_{2j}(x-\xi_{2j})_+.
$$
Now we can construct the special $\Relu$ network $\Up^{4,2L}$ that generates T via the successive transformations
$A^{(j)}$ given by the matrices
$$
M^{(1)}=\begin{bmatrix}1&1&1&0\end{bmatrix}^T, \quad
b^{(1)}=\begin{bmatrix}0&-\xi_1&-\xi_2&0\end{bmatrix}^T,
$$
The $j$th layer, $j=2,\ldots,2L$,  produces $S_{j-1}$  in its CC node
via the matrix
$$
M^{(j)}=\begin{bmatrix}1&0&0&0 \\
                                    1&0&0&0 \\
                                    1&0&0&0 \\
0&c_{2j-3}&c_{2j-2}&1 
\end{bmatrix}, \quad 
b^{(j)}=\begin{bmatrix}0\\-\xi_{2j-1}\\-\xi_{2j}\\0\end{bmatrix}.
$$
Finally, the output layer  is given by the matrix
$$
M^{(2L)}=\begin{bmatrix}a&c_{2L-1}&c_{2L}&1\end{bmatrix},\quad b^{(2L)}=b,
$$
where the first entry $a$ and the bias $b$ account for the linear function $ax+b$ in $T$. 
In this case we have 
$\Sigma_{4L}\subset \Up^{4,2L}\subset \Upsilon^{4,2L}$, 
with  number of parameters 
$$
n(4,2L)=40L-7=40\left\lceil \frac{n}{4}\right\rceil-7<10n+33<19n, \quad n\geq 4.
$$
For the case $n<4$, we again add artificial breakpoints so that we represent 
$T\in \Sigma_n\subset \Sigma_{4}$  as
$$
T(x)=ax+b+\sum_{j=1}^{4}c_j(x-\xi_j)_+=ax+b+\sum_{j=1}^{2}S_j, \quad 
S_j:=c_{2j-1}(x-\xi_{2j-1})_++c_{2j}(x-\xi_{2j})_+,
$$
and as above generate a special $\Relu$ network $\Up^{4,2}$ for which $\Sigma_n\subset \Up^{4,2}$, and whose 
parameters 
$$
n(4,2)=33=W^2+4W+1, \quad W=4.
$$

Now, for the case $(W-2)\in\{3,4,5\}$, let us first consider $n\geq 2(W-2)$ and take $L:=\left\lceil\frac{n}{2(W-2)}\right\rceil$. 
If $n<2(W-2)L$, we add artificial breakpoints so that we represent 
$T\in \Sigma_n\subset \Sigma_{2(W-2)L}$.
We do the same construction as in the case $W-2=2$, by dividing the indices $\{1,\ldots,2(W-2)L\}$ into $2L$ groups of $W-2$ numbers, as shown in
$$
T(x)=ax+b+\sum_{j=1}^{2(W-2)L}c_j(x-\xi_j)_+=ax+b+\sum_{j=1}^{2L}S_j, \quad 
S_j:=\sum_{i=0}^{W-3}c_{(W-2)j-i}(x-\xi_{(W-2)j-i})_+,
$$
and execute the same construction as before by concatenating the networks producing $S_j$.
In this case, we have 
$\Sigma_n\subset \Sigma_{2(W-2)L}\subset \Up^{W,2L}$, 
and when $n\geq 2(W-2)$,
$$
n(W,2L)=2W(W+1)\left\lceil\frac{n}{2(W-2)}\right\rceil-(W-1)^2+2<\frac{W(W+1)}{W-2}n+W^2+4W+1<25n.
$$
When $n<2(W-2)$, we again add artificial breakpoints so that we represent 
$T\in \Sigma_n\subset \Sigma_{2(W-2)}$  as
$$
T(x)=ax+b+\sum_{j=1}^{2(W-2)}c_j(x-\xi_j)_+=ax+b+\sum_{j=1}^{2}S_j, \quad 
S_j:=\sum_{i=0}^{W-3}c_{(W-2)j-i}(x-\xi_{(W-2)j-i})_+,
$$
and as above generate a special $\Relu$ network $\Up^{W,2}$ with depth $L=2$ for which $\Sigma_n\subset \Up^{W,2}$, and whose 
parameters 
$$
n(W,2)=2W(W+1)-(W-1)^2+2=W^2+4W+1, \quad n<2(W-2).
$$
This completes the proof. \hfill $\square$

\subsection{Proof of Theorem \ref{compos} }
\label{ss:structure}

{\bf Proof:}
Note that for every $k$-tuple $(\tilde S_k,\cdots,\tilde S_1)\in \Sigma_{n_k}\times\cdots\times\Sigma_{n_1}$,
 we can find another  $k$-tuple $(S_k,\ldots,S_1)\in \Sigma_{n_k}\times\cdots\times\Sigma_{n_1}$, which we call a representative of the composition, with the properties:
 \begin{itemize}
\item $S_j([0,1])\subset [0,1]$, $j=1,\ldots,k-1$.
\item $\tilde S_k\circ\cdots\circ \tilde S_1=S_k\circ\cdots\circ S_1$.
\end{itemize}
Indeed, if we denote by $m_1:=\min_{x\in [0,1]}\tilde S_1(x)$, $M_1:=\max_{x\in [0,1]}\tilde S_1(x)$,   
define inductively
$$
m_j:=\min_{x\in[m_{j-1},M_{j-1}]}\tilde S_j, \quad M_j:=\max_{x\in[m_{j-1},M_{j-1}]}\tilde S_j, \quad j=2,\ldots,k-1,
$$
and consider the functions 
\begin{eqnarray}
\nonumber
S_1&:=&\frac{\tilde S_1-m_1}{M_1-m_1}\in \Sigma_{n_1},\\
\nonumber
S_j&:=&\frac{\tilde S_j(x(M_{j-1}-m_{j-1})+m_{j-1})-m_{j-1}}{M_j-m_j}\in \Sigma_{n_j}, \quad j=2,\ldots,k-1,\\
\nonumber
S_k&:=&\tilde S_k(x(M_{k-1}-m_{j-1})+m_{k-1}).
\end{eqnarray}
The $k$-tuple  $(S_k,\ldots,S_1)$ will be  a representative of the  composition $\tilde S_k\circ\ldots\circ \tilde S_1$. So, in going further, we will always assume that we are dealing with  representatives of all compositions we consider and with $\Relu$ networks that output these representatives. 

 Relation  \eqref{E:pure-compos} follows from Proposition \ref{composnetsup} and Theorem \ref{T:main}.  Indeed, 
if we fix an element in 
$\Sigma^{n_k\circ\cdots\circ n_1}:=\{\tilde S_k\circ\cdots\circ \tilde S_1:\,\tilde S_j\in \Sigma_{n_j}, j=1,\ldots,k\}$ and consider its representative $(S_k,\ldots,S_1)$, each $S_j$ in the composition $S_k\circ\cdots\circ S_1$ can be produced by a $\Relu$ network ${\cal C}_j$ with width $W$ and depth 
$$
L_j=2\left \lceil \frac{n_j}{\lfloor \frac{W-2}{6}\rfloor(W-2)}\right\rceil, 
$$
 and therefore, part {\bf (i)} of Proposition \ref{composnetsup} ensures that 
$S_k\circ\cdots\circ S_1\in \Upsilon^{W,\sum_{j=1}^kL_j}$.
A similar estimate as in the proof of  Theorem \ref{T:main} yields
\begin{eqnarray}
\nonumber
n(W,L)
&<&34\sum_{j=1}^kn_j+2k(W^2+W), 
\end{eqnarray}
as desired. 

To establish \eqref{E:full-compos}, for each $i=1,\ldots,m$, let us denote by ${\cal N}_i$ the $\Relu$ network from 
\eqref{E:pure-compos}  with width $W-2$ that produces the composition $S_{i,\ell_i}\circ\cdots\circ S_{i,1}$ and has depth 
$$
L_i=L(n_{i,\ell_i},\ldots,n_{i,1})=2\sum_{j=1}^{\ell_i}\left \lceil \frac{n_{i,j}}{\lfloor \frac{W-4}{6}\rfloor
(W-4)}\right\rceil.
$$
Then, Proposition \ref{composnetsup}, part {\bf (ii)}  gives
$$
S=\sum_{i=1}^ma_iS_{i,\ell_i}\circ\cdots\circ S_{i,1} \in \Up^{W,L},
$$
with
\begin{eqnarray}
\nonumber
L=\sum_{i=1}^mL_i=2\sum_{i=1}^m\sum_{j=1}^{\ell_i}\left \lceil \frac{n_{i,j}}{\lfloor \frac{W-4}{6}\rfloor(W-4)}\right\rceil.
\end{eqnarray}
A similar estimate as in the proof of  Theorem \ref{T:main} yields
\begin{eqnarray}
\nonumber
n(W,L)
&<&44\sum_{i=1}^m\sum_{j=1}^{\ell_i}n_{i,j}+2W(W+1)\sum_{i=1}^m\ell_i. \nonumber
\end{eqnarray}

As discussed in Remark \ref{relufree}, $\Up^{W,L}$ can always be viewed as a subset of $\Upsilon^{W,L}$, 
and the proof is completed.
\hfill$\Box$

\subsection{Proof of Proposition \ref{PropRiesz}}

Let us first start with the  notation
\begin{eqnarray}
\nonumber
{\bf 1}_{\{i=j \}}:=
\begin{cases}
    1, \quad & i=j, \\
    0, & i\neq j,
  \end{cases}
\end{eqnarray}
and isolate the following technical observation.

\begin{lemma}
\label{LemSum}
For any nonnegative sequence $u \in \ell_2(\N)$,
\begin{equation}
\label{OblFrame}
\sum_{\substack{k,\ell \ge 1\\ k \not= \ell} }
u_k u_\ell \sum_{m,n\ge 0} \frac{1}{(2m+1)^2} \frac{1}{(2n+1)^2} {\bf 1}_{\{(2m+1)k = (2n+1)\ell \}}
\le \frac{\pi^4}{192} \|u\|_2^2.
\end{equation}
\end{lemma}

\noindent
 {\bf Proof}.
 For each integer $m \ge 0$, let us introduce the sequence $u^{(2m+1)} \in \ell_2(\N)$ defined by
$$
u^{(2m+1)}_j = \left\{
\begin{matrix}
u_{\frac{j}{2m+1}}, & \mbox{if } j \in (2m+1)\N,\\
0, & \mbox{if } j \not\in (2m+1)\N,
\end{matrix}
\right.
$$
i.e.,  we  consider a new sequence obtained from the original one by separating every two consecutive
terms with  $2m$ zeroes, starting with $2m$ zeroes. 
We easily see that
\begin{eqnarray*}
\langle u^{(2m+1)}, u^{(2n+1)} \rangle
 &=& \sum_{j \in \N} u^{(2m+1)}_j u^{(2n+1)}_j
%=  \sum_{j \in (2m+1)\N} u^{(2m+1)}_j u^{(2n+1)}_j\\
%&=& \sum_{k \in \N} u_k u_{\frac{2m+1}{2n+1}k} {\bf 1}_{\{ (2m+1)k = (2n+1)\ell \mbox{ for some } \ell \in \N \}}\\
%&=& 
=\sum_{k,\ell \in \N} u_k u_\ell {\bf 1}_{\{ (2m+1)k = (2n+1)\ell\}},
\end{eqnarray*}
and in particular $\|u^{(2m+1)}\|_2^2=\|u\|_2^2$ for every $m \ge 0$.
Thus, the left-hand side of \eqref{OblFrame}, which we denote by $\Sigma$, can be written as
 \begin{eqnarray}
 \nonumber
 \Sigma & = & \sum_{\substack{m,n\ge 0\\ m \not= n}} \frac{1}{(2m+1)^2} \frac{1}{(2n+1)^2}
 \sum_{k,\ell \ge 1 }
u_k u_\ell  {\bf 1}_{\{(2m+1)k = (2n+1)\ell \}}\\
\nonumber
& = & \sum_{\substack{m,n\ge 0\\ m \not= n}} \frac{1}{(2m+1)^2} \frac{1}{(2n+1)^2} \langle u^{(2m+1)}, u^{(2n+1)} \rangle\\
\label{Sigma2Parts}
&=& \bigg\| \sum_{m \ge 0} \frac{1}{(2m+1)^2} u^{(2m+1)} \bigg\|_2^2
- \sum_{m \ge 0} \frac{1}{(2m+1)^4} \|u\|_2^2.
 \end{eqnarray}
 By a simple triangle inequality, we have
 \be
 \label{SigmaPart1}
 \bigg\| \sum_{m \ge 0} \frac{1}{(2m+1)^2} u^{(2m+1)} \bigg\|_2 
 \le \sum_{m \ge 0} \frac{1}{(2m+1)^2}  \|u\|_2
 = \frac{\pi^2}{8} \|u\|_2.
 \ee
Moreover, it is well-known that  
\be
\label{SigmaPart2}
\sum_{m \ge 0} \frac{1}{(2m+1)^4} = \frac{\pi^4}{96}.
\ee 
Substituting \eqref{SigmaPart1} and \eqref{SigmaPart2} into \eqref{Sigma2Parts} yields the announced result. 
 \hfill $\Box$

 \noindent
 {\bf Proof of Proposition \ref{PropRiesz}}.
 We equivalently prove the result for the $L_2$-normalized version of the system $(\mathcal{C}_k,\mathcal{S}_k)_{k \ge 1}$,
 i.e., for $(\wt{\mathcal{C}}_k := \sqrt{3} \ \mathcal{C}_k,\wt{\mathcal{S}}_k := \sqrt{3} \ \mathcal{S}_k)_{k \ge 1}$.
 Let $(c_k,s_k)_{k \ge 1}$ denote the orthonormal basis for $L_2^0[0,1]$
 made of the usual trigonometric functions
  $$
 c_k(x) = \sqrt{2} \cos(2 \pi k x),
 \qquad
 s_k(x) = \sqrt{2} \sin(2 \pi k x), \qquad x \in [0,1].
 $$
It is routine to verify (by computing Fourier series) that
$$
{\cal C} = \lambda \sum_{m \ge 0} \frac{1}{(2m+1)^2} c_{2m+1},
\qquad
{\cal S} = \lambda \sum_{m \ge 0} \frac{(-1)^m}{(2m+1)^2} s_{2m+1},
$$
for some constant $\lambda >0$, 
from which one immediately obtains
that, for any $k \ge 1$,
$$
\wt{\cal C}_k = \mu \sum_{m \ge 0} \frac{1}{(2m+1)^2} c_{(2m+1)k},
\qquad
\wt{\cal S}_k = \mu \sum_{m \ge 0} \frac{(-1)^m}{(2m+1)^2} s_{(2m+1)k},
$$ 
for some constant $\mu >0$.
The normalization $\|{\cal C}\|_{L_2[0,1]} = \|{\cal S}\|_{L_2[0,1]} = 1$ imposes
$$
\mu^2 \sum_{m \ge 0} \frac{1}{(2m+1)^4} = 1,
\qquad \mbox{i.e.,}
\qquad
\mu^2 \frac{\pi^4}{96}=1. 
$$ 
Let us introduce operators $T_\mathcal{C}, T_{\mathcal{S}}$
defined for $v \in \ell_2(\N)$ and $j \in \N$, by
\begin{eqnarray*}
T_{\mathcal{C}}(v)_j
& = & \sum_{k \ge 1} v_k \langle \wt{\mathcal{C}}_k,c_j \rangle 
= \mu \sum_{k \ge 1} v_k \sum_{m \ge 0} \frac{1}{(2m+1)^2} {\bf 1}_{ \{ (2m+1)k = j \} },
\\
T_{\mathcal{S}}(v)_j
& = & \sum_{k \ge 1} v_k \langle \wt{\mathcal{S}}_k,s_j \rangle 
= \mu \sum_{k \ge 1} v_k \sum_{m \ge 0} \frac{(-1)^m}{(2m+1)^2} {\bf 1}_{ \{ (2m+1)k = j \} },
\end{eqnarray*}
and let us first verify that these are well-defined operators from $\ell_2(\N)$ to $\ell_2(\N)$,
i.e., that both $\|T_{\mathcal{C}}v\|_2$ and  $\|T_{\mathcal{S}}v\|_2$ are finite when $v \in \ell_2(\N)$.
To do so, we observe that
\begin{eqnarray*}
\|T_{\mathcal{C}} v\|_2^2 
& = &  
\mu^2 \sum_{j \ge 1} 
\sum_{k,\ell \ge 1} v_k v_\ell \sum_{m,n \ge 0} \frac{1}{(2m+1)^2} \frac{1}{(2n+1)^2} {\bf 1}_{ \{ (2m+1)k = j \} } {\bf 1}_{ \{ (2n+1)\ell = j \} }\\
& = & 
\Sigma_{(=)} + \Sigma_{(\not=)},
\end{eqnarray*}
where $\Sigma_{(=)}$ represents the contribution to the sum when $k$ and $\ell$ are equal
and $\Sigma_{(\not=)}$ represents the contribution to the sum when $k$ and $\ell$ are distinct.
We notice that 
$$
\Sigma_{(=)}
= \sum_{k\ge 1} v_k^2 \, 
\mu^2
 \sum_{m\ge 0} \frac{1}{(2m+1)^4}   \sum_{j \ge 1}  {\bf 1}_{ \{ (2m+1)k = j \} } 
 = 
 \sum_{k\ge 1} v_k^2 \,
\mu^2
 \sum_{m\ge 0} \frac{1}{(2m+1)^4} 
  =\sum_{k\ge 1} v_k^2. 
$$
Therefore, relying on Lemma \ref{LemSum}, we obtain
\begin{eqnarray}
\nonumber
\left| \|T_{\mathcal{C}} v\|_2^2  - \|v\|_2^2 \right|
& = & \left| \Sigma_{(\not=)} \right|
\le \mu^2 
\sum_{\substack{k,\ell \ge 1\\ k \not= \ell}} |v_k| |v_\ell | \sum_{m,n \ge 0} \frac{1}{(2m+1)^2} \frac{1}{(2n+1)^2} \sum_{j \ge 1}  {\bf 1}_{ \{ (2m+1)k = j \} } {\bf 1}_{ \{ (2n+1)\ell = j \} }\\
\nonumber
& = & 
 \mu^2 
\sum_{\substack{k,\ell \ge 1\\ k \not= \ell}} |v_k| |v_\ell | \sum_{m,n \ge 0} \frac{1}{(2m+1)^2} \frac{1}{(2n+1)^2}   {\bf 1}_{ \{ (2m+1)k =  (2n+1)\ell  \} } \\
\label{BoundSigmanot=}
& \le & \mu^2 \frac{\pi^4}{192} \|v\|_2^2 = \frac{1}{2} \|v\|_2^2.
\end{eqnarray}
This clearly justifies that $\|T_{\mathcal{C}} v\|_2^2 < \infty$,
and $\|T_{\mathcal{S}} v\|_2^2 < \infty$ is verified in a similar fashion.
In fact, the inequality \eqref{BoundSigmanot=} and the analogous one for $T_{\mathcal{S}}$ show that
\be
\label{T*T-I}
\|T_{\mathcal{C}}^* T_{\mathcal{C}} - I \|_{2 \to 2}
= \max_{\|v\|_2 = 1} | \langle v, (T_{\mathcal{C}}^* T_{\mathcal{C}} - I) v \rangle  |
\le \frac{1}{2},
\qquad 
\|T_{\mathcal{S}}^* T_{\mathcal{S}} - I \|_{2 \to 2} \le \frac{1}{2}.
\ee
This ensures that the operators $T_{\mathcal{C}}^* T_{\mathcal{C}}$ and $T_{\mathcal{S}}^* T_{\mathcal{S}}$ are invertible.
Let us admit for a while that  the operators $T_{\mathcal{C}} T_{\mathcal{C}}^*$ and $T_{\mathcal{S}} T_{\mathcal{S}}^*$ are also invertible.
Then we derive that $T_{\mathcal C}$ is invertible with inverse $(T_{\mathcal C}^* T_{\mathcal C})^{-1} T_{\mathcal C}^*$,
since $(T_{\mathcal C}^* T_{\mathcal C})^{-1} T_{\mathcal C}^* T_{\mathcal C} = I $ is obvious 
and $T_{\mathcal{C}} (T_{\mathcal C}^* T_{\mathcal C})^{-1} T_{\mathcal C}^* = I$
is equivalent, by the invertibility of $T_{\mathcal C} T_{\mathcal C}^*$, to 
$T_{\mathcal C} T_{\mathcal C}^* T_{\mathcal{C}} (T_{\mathcal C}^* T_{\mathcal C})^{-1} T_{\mathcal C}^* = T_{\mathcal C} T_{\mathcal C}^*$,
which is obvious.
We derive that $T_{\mathcal{S}}$ is invertible in a similar fashion.
From here, we can show that the system $(\wt{\mathcal{C}}_k, \wt{\mathcal{S}}_k)_{k \ge 1}$
spans $L_{2}^0[0,1]$.
Indeed, we claim that any $f \in L_{2}^0[0,1]$ can be written,
with $\alpha:= (\langle f, c_j \rangle)_{j \ge 1}$ and $\beta:= (\langle f, s_j \rangle)_{j \ge 1}$, as
$$
f = \sum_{k \ge 1} (T_{\mathcal{C}}^{-1} \alpha)_k \wt{\mathcal{C}}_k
+ \sum_{k \ge 1} (T_{\mathcal{S}}^{-1} \beta)_k \wt{\mathcal{S}}_k.
$$ 
This identity is verified by taking the inner product with any $c_j$ and any $s_j$.
For instance, the right-hand side has an inner product with $c_j$ equal to
$$
\sum_{k \ge 1} (T_{\mathcal{C}}^{-1} \alpha)_k \langle \wt{\mathcal{C}}_k, c_j \rangle + 0
= \left(T_{\mathcal{C}} (T_{\mathcal{C}}^{-1} \alpha)\right)_j = \alpha_j = \langle f, c_j \rangle,
$$
which  confirms our claim.
As for a normalized version of \eqref{Frame}, it follows from \eqref{T*T-I} by noticing that 
$$
\bigg\| \sum_{k\ge 1} (a_k \wt{\cal C}_k + b_k \wt{\cal S}_k)
\bigg\|_{L_2[0,1]}^2 - (\| a \|_2^2 + \| b \|_2^2)
= \bigg\|\sum_{k\ge 1} a_k \wt{\cal C}_k 
\bigg\|_{L_2[0,1]}^2 - \| a \|_2^2 
+ \bigg\| \sum_{k\ge 1}  b_k \wt{\cal S}_k
\bigg\|_{L_2[0,1]}^2 - \|b \|_2^2,
$$
combined with the observation that
\begin{eqnarray*}
\bigg\| \sum_{k\ge 1} a_k \wt{\cal C}_k 
\bigg\|_{L_2[0,1]}^2 - \| a \|_2^2 
& = & \sum_{j \ge 1}  \bigg( \sum_{k \ge 1} a_k \left\langle\wt{\mathcal{C}}_k , c_j \right\rangle\bigg)^2 - \|a\|_2^2
= \sum_{j \ge 1} (T_{\mathcal{C}}a)_j^2 - \|a\|_2^2 = \|T_{\mathcal{C}} a\|_2^2 - \|a\|_2^2\\
& = & \langle (T_{\mathcal{C}}^* T_{\mathcal{C}} -I) a, a \rangle 
\le \frac{1}{2} \|a\|_2^2,
\end{eqnarray*}
and the similar observation that
$$
\bigg\| \sum_{k\ge 1} b_k \wt{\cal S}_k 
\bigg\|_{L_2[0,1]}^2 - \| b \|_2^2
\le \frac{1}{2} \|b\|_2^2.
$$
We deduce that a normalized version of \eqref{Frame} holds with constants $\wt{c}=1/2$ and $\wt{C}=3/2$,
hence \eqref{Frame} holds with $c=1/6$ and $C=1/2$.

It now remains to establish that the operators $T_{\mathcal{C}} T_{\mathcal{C}}^*$ and 
$T_{\mathcal{S}} T_{\mathcal{S}}^*$ are invertible, which we do by showing that
\be
\label{TT*-I}
\| T_{\mathcal{C}} T_{\mathcal{C}}^* - I \|_{2 \to 2} \le \rho
\qquad \mbox{and} \qquad
\| T_{\mathcal{S}} T_{\mathcal{S}}^* - I \|_{2 \to 2} \le \rho
\ee
for some constant $\rho < 1$.
We concentrate on the case of $T_{\mathcal{C}}$,
as the case of $T_{\mathcal{S}}$ is handled similarly.
We first remark that the adjoint of $T_{\mathcal{C}}$ is given,
for any $v \in \ell_2(\N)$ and $j \in \N$, by
$$
T_{\mathcal{C}}^*(v)_j
= \sum_{k \ge 1} v_k \langle \wt{\mathcal{C}}_j,c_k \rangle 
= \mu \sum_{k \ge 1} v_k \sum_{m \ge 0} \frac{1}{(2m+1)^2} {\bf 1}_{ \{ (2m+1)j = k \} }.
$$
We then compute
\begin{eqnarray*}
\|T_{\mathcal{C}}^* v\|_2^2 
& = &  
\mu^2 \sum_{j \ge 1} 
\sum_{k,\ell \ge 1} v_k v_\ell \sum_{m,n \ge 0} \frac{1}{(2m+1)^2} \frac{1}{(2n+1)^2} {\bf 1}_{ \{ (2m+1)j = k \} } {\bf 1}_{ \{ (2n+1)j = \ell \} }\\
& = & 
\Sigma_{(=)}^* + \Sigma_{(\not=)}^*,
\end{eqnarray*}
where $\Sigma_{(=)}^*$ represents the contribution to the sum when $k$ and $\ell$ are equal
and $\Sigma_{(\not=)}^*$ represents the contribution to the sum when $k$ and $\ell$ are distinct.
We notice that 
$$
\Sigma_{(=)}^* = 
\sum_{k \ge 1} v_k^2 \, \mu^2  \sum_{m \ge 0} \frac{1}{(2m+1)^4} \sum_{j \ge 1}  {\bf 1}_{ \{ (2m+1)j = k \} } 
$$ 
satisfies, on the one hand,
$$
\Sigma_{(=)}^* \le 
\sum_{k \ge 1} v_k^2 \, \mu^2  \sum_{m \ge 0} \frac{1}{(2m+1)^4}
= \sum_{k \ge 1} v_k^2 = \|v\|_2^2,
$$
and on the other hand, by considering only the summand for $m=0$ and $j=k$,
$$
\Sigma_{(=)}^* \ge 
\sum_{k \ge 1} v_k^2 \, \mu^2 = \mu^2 \|v\|_2^2.
$$
Moreover, we have 
\begin{eqnarray}
\nonumber
 \left| \Sigma_{(\not=)} \right|
 & \le &
 \mu^2 
\sum_{\substack{k,\ell \ge 1\\ k \not= \ell}} |v_k| |v_\ell | \sum_{m,n \ge 0} \frac{1}{(2m+1)^2} \frac{1}{(2n+1)^2} \sum_{j \ge 1}  {\bf 1}_{ \{ (2m+1)j = k \} } {\bf 1}_{ \{ (2n+1)j = \ell \} }\\
\nonumber
& \le & 
 \mu^2 
\sum_{\substack{k,\ell \ge 1\\ k \not= \ell}} |v_k| |v_\ell | \sum_{m,n \ge 0} \frac{1}{(2m+1)^2} \frac{1}{(2n+1)^2}   {\bf 1}_{ \{ (2m+1)\ell =  (2n+1)k  \} } \\
& \le & \mu^2 \frac{\pi^4}{192} \|v\|_2^2 = \frac{1}{2} \|v\|_2^2,
\end{eqnarray}
where the last inequality used Lemma \ref{LemSum} again.
Therefore, we obtain
$$
\left| \langle (T_{\mathcal{C}} T_{\mathcal{C}}^* - I)v, v \rangle \right|
= \left| \|T_{\mathcal{C}}^*v\|_2^2 - \|v\|_2^2 \right|
= \left| (\Sigma_{(=)}^* - \|v\|_2^2) + \Sigma_{(\not=)}^* \right|
\le (1-\mu^2) \|v\|_2^2 + \frac{1}{2} \|v\|_2^2.
$$
Taking the maximum over all $v \in \ell_2(\N)$ with $\|v\|_2=1$,
we arrive at the result announced in \eqref{TT*-I} with $\rho := 1-\mu^2 + 1/2 \le 0.5145$. The proof is now complete.
 \hfill $\Box$

 \vskip .2in
 \noindent
 ID, Dept. of Mathematics, Duke University, Durham, NC 27708; ingrid@math.duke.edu
 
 \noindent
\{RD,SF,BH,GP\},  Dept. of Mathematics,  Texas A$\&$M University,  College Station, TX, 77843; \{rdevore,foucart,bhanin,gpetrova\}@math.tamu.edu

\noindent
BH,  Facebook AI Research, NYC; bhanin@fb.com

 \end{document}

%% file: NN-macros.tex
\def\RR{\rm \hbox{I\kern-.2em\hbox{R}}}
\def\NN{\rm \hbox{I\kern-.2em\hbox{N}}}
\def\ZZ{\rm {{\rm Z}\kern-.28em{\rm Z}}}
\def\CC{\rm \hbox{C\kern -.5em {\raise .32ex \hbox{$\scriptscriptstyle
|$}}\kern
-.22em{\raise .6ex \hbox{$\scriptscriptstyle |$}}\kern .4em}}

\def\<{\langle}
\def\>{\rangle}

\def\wt{\widetilde}

\def\cS{{\cal S}}

\def\cA{{\cal A}}

\def\cN{{\cal N}}

\def\cM{{\cal M}}

\def\cP{{\cal P}}

\def\cP{{\cal P}}

\def\R{\mathbb{R}}
\def\N{\mathbb{N}}

\def\Relu{\mathrm{ReLU}}
\def\Chi{\raise .3ex
\hbox{\large $\chi$}} 
\def\lsima{\hbox{\kern -.6em\raisebox{-1ex}{$~\stackrel{\textstyle<}{\sim}~$}}\kern -.4em}
\def\lsim{\hbox{\kern -.2em\raisebox{-1ex}{$~\stackrel{\textstyle<}{\sim}~$}}\kern -.2em}
\def\gsim{\hbox{\kern -.2em\raisebox{-1ex}{$~\stackrel{\textstyle>}{\sim}~$}}\kern -.2em}
\def\[{\Bigl [}
\def\]{\Bigr ]}
\def\({\Bigl (}
\def\){\Bigr )}
\def\[{\Bigl [}
\def\]{\Bigr ]}
\def\({\Bigl (}
\def\){\Bigr )}

\newcommand{\be}{\begin{equation}}
\newcommand{\ee}{\end{equation}}
\newcommand{\bea}{$$ \begin{array}{lll}}
\newcommand{\eea}{\end{array} $$}
\newcommand{\bi}{\begin{itemize}}
\newcommand{\ei}{\end{itemize}}
\newcommand{\iref}[1]{(\ref{#1})}
\newcommand{\eref}[1]{\eqref{#1}}
\newtheorem{theorem}{Theorem}[section]     % changed by Simon
\newtheorem{remark}{Remark}[section]         % changed by Simon
\newtheorem{lemma}[theorem]{Lemma}
\newtheorem{proposition}[theorem]{Proposition} 

\newtheorem{definition}{Definition}[section]  % changed by Simon

\def\Up{\overline{\underline{\Upsilon}}}
\newlength{\Sigmalength}
\newlength{\Upsilonlength}
\def\N{\mathbb N}

\newcommand{\set}[1]{\left\{#1\right\} }
\newcommand{\gives}{\rightarrow}